\documentclass[twoside,journal]{IEEEtran}

\usepackage{amsmath,bm,nccmath,gensymb}

\usepackage{graphicx,stfloats,scalerel}
\usepackage{flushend}

\usepackage{cite,url,algorithmic,array,multirow,dsfont,balance,enumitem,tabularx,booktabs}

\usepackage[svgnames]{xcolor}
\usepackage[bookmarks=false,linkcolor=Navy,urlcolor=Navy,colorlinks,citecolor=Navy]{hyperref}

\fnbelowfloat
\begin{document}
\bstctlcite{IEEEexample:BSTcontrol}

\title{Equivariant Wavelets: Fast Rotation and Translation Invariant Wavelet Scattering Transforms}

\author{Andrew~K.~Saydjari \href{https://orcid.org/0000-0002-6561-9002}{\includegraphics[scale=0.03,trim=1cm -2cm 0 10cm]{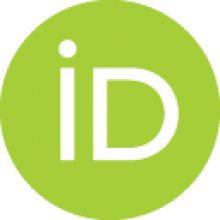}} and Douglas~P.~Finkbeiner\href{https://orcid.org/0000-0003-2808-275X}{\includegraphics[scale=0.03,trim=-3cm -2cm 0 10cm]{figures/Orcid-ID.png}}%
\thanks{A. K. Saydjari and D. P. Finkbeiner are with the Department of Physics, Harvard University, 17 Oxford St., Cambridge, MA 02138, USA and the Harvard-Smithsonian Center for Astrophysics, 60 Garden St., Cambridge, MA 02138, USA (e-mail: andrew.saydjari@cfa.harvard.edu, dfinkbeiner@cfa.harvard.edu).}%
\thanks{Manuscript received X X, 2021; revised X X, 2021.}
\thanks{ }}

\markboth{IEEE Transactions on Signal Processing,~Vol.~X, No.~X, X~2021}%
{Saydjari and Finkbeiner: Equivariant Wavelets}

\maketitle


\begin{abstract}
Wavelet scattering networks, which are convolutional neural networks (CNNs) with fixed filters and weights, are promising tools for image analysis. Imposing symmetry on image statistics can improve human interpretability, aid in generalization, and provide dimension reduction. In this work, we introduce a fast-to-compute,  translationally  invariant  and rotationally equivariant wavelet scattering network (EqWS) and filter bank of wavelets (triglets). We demonstrate the interpretability and quantify the invariance/equivariance of the coefficients, briefly commenting on difficulties with implementing scale equivariance. On MNIST, we show that training on a rotationally invariant reduction of the coefficients maintains rotational invariance when generalized to test data and visualize residual symmetry breaking terms. Rotation equivariance is leveraged to estimate the rotation angle of digits and reconstruct the full rotation dependence of each coefficient from a single angle. We benchmark EqWS with linear classifiers on EMNIST and CIFAR-10/100, introducing a new second-order, cross-color channel coupling for the color images. We conclude by comparing the performance of an isotropic reduction of the scattering coefficients and RWST, a previous coefficient reduction, on an isotropic classification of magnetohydrodynamic simulations with astrophysical relevance.

\end{abstract}



\section{Introduction}
\IEEEPARstart{P}{hysical} models which are simple and valid for a large class of problems often gain the most traction in applications. However, simplicity that might improve interpretability also often limits how generic a model can be. For statistical analysis of 2D-fields (i.e., images), an extremely simple descriptor is the 2-point correlation function (2PCF), or its Fourier space analogue, the power spectrum. While the Gaussian process assumption implicit in only considering the power spectrum has great success for some applications, such as analysis of the Cosmic Microwave Background (CMB) \cite{PlanckCollaboration:2017:A&A:}, information on phase coherent structures, such as filaments, is lost \cite{Peek:2019:ApJL:}.

Higher-order correlation functions, such as the 3-point correlation function (3PCF) and its Fourier-domain analogue, the bispectrum, provide an improvement in capturing higher-order correlations \cite{Peebles:2001:ASPC:}, but interpretation of these higher order statistics remains difficult \cite{Burkhart:2009:ApJ:, Burkhart:2016:ApJ:}. Convolutional neural networks (CNNs) capture higher-order correlations through successive convolutions and non-linearities and achieve state-of-the-art performance on image analysis tasks \cite{zeiler2014visualizing}. While some progress has been made in feature interpretation, CNNs are similarly difficult to interpret \cite{zhou2014object, zeiler2014visualizing, karpathy2015visualizing, olah2017feature, bau2017network, cammarata:2020:curve}.

We will focus here on wavelet scattering networks, which have an intermediate complexity, using fixed convolutional filters and a simple modulus nonlinearity \cite{Bruna:2012:arXiv:,Mallat:2011:arXiv:}. After a small modification, the first order wavelet scattering coefficients can be interpreted as sampling the power spectrum. These networks have had success in texture classification \cite{Bruna:2012:arXiv:}, molecular structure calculation \cite{Hirn:2016:arXiv:,Eickenberg:2018:JChPh:}, turbulence classification \cite{Bruna:2013:arXiv:,Allys:2019:A&A:,saydjari2020classification,Regaldo-SaintBlancard:2020:arXiv:}, and cosmological parameter inference \cite{Allys:2020:arXiv:,villaescusa2020quijote,Cheng:2020:arXiv:,Cheng:2021:arXiv:}. In addition, progress has been made on image reconstruction and denoising from wave scattering transform (WST) coefficients using generative networks \cite{Angles:2018:arXiv:} and gradient descent \cite{Bruna:2018:arXiv:,Regaldo-SaintBlancard:2021:arXiv:}.

Symmetry drives simplicity in physical models. Without symmetry, searching for the laws of physics would be akin to trying to find patterns in correlations in an extremely high-dimensional phase space. Thus, when building data analysis tools in physics, those tools should encode and respect the symmetries of the problem. This allows the analysis to make the best use of available data toward constraining the model behavior along free (unconstrained by symmetry) dimensions. Two common symmetries of interest are translation and rotation.

Machine learning algorithms often struggle to learn symmetries of a problem, because they are trained on finite, noisy data, while symmetries are exact. Thus, we must incorporate prior knowledge of the symmetry of the problem into the algorithm design. For example, if the position of an object in an image has no bearing on its class, we want the image classification algorithm to be invariant under image translations.

However, the workhorse of machine learning image analysis, the convolutional neural network (CNN) is generally not translation invariant \cite{engstrom2019exploring}. The lack of translation invariance derives from the common practice of down-sampling via strided-convolution or (max/average)-pooling \cite{scherer2010evaluation,azulay2018deep,zhang2019making}. However, this down-sampling is necessary for computational feasibility of modern networks on large images. In this case, the CNN becomes only translationally invariant to translations that are multiples of the down-sampling period \cite{zhang2019making}. Many works have explored how to suppress the magnitude of these effects, such as anti-aliasing (low-pass) filtering \cite{lecun1998gradient,zhang2019making}. This filtering forces the network to obey the Shannon-Nyquist sampling theorem by eliminating frequencies above the down-sampling frequency, a procedure which inherently involves information loss \cite{nyquist1928certain}.

In contrast to CNNs, wavelet-based techniques do not need to down-sample the convolution and thus can avoid breaking the exact translation equivariance inherent in the Fourier transform involved in convolutions.\footnote{While wavelet-based techniques do not need to use down-sampling and low-pass filtering, implementations often do. To our knowledge, the code we release here is the first public wavelet scattering code to not use sub-sampling and be exactly translationally invariant.} This is possible because in wavelet-based techniques, the convolutional filters are pre-specified, not learned. It is no more difficult to specify a filter the size of the entire image, but it is much more difficult to perform pixel-wise gradient descent on large filters.\footnote{While we focus on the wavelet scattering transform, much of the discussion with respect to wavelet or filter engineering has implications for the general set of methods which use kernels (pre-specified filters), such as structured receptive field networks \cite{jacobsen2016structured}.}

In building a model which respects the symmetry of a problem, invariance can be too restrictive a requirement. For example, if all filters were rotationally symmetric, there would be no way of describing the relative orientation of two objects. In an equivariant representation, the set of coefficients representing an image before and after an image rotation will be related by a corresponding operation on that coefficient space. If we then want to solve a rotationally invariant problem, we can construct combinations of the equivariant coefficients to make a rotationally invariant set of coefficients, which depend only on the relative angles between objects in the image but not on absolute angles.

There exists a large body of work on rotational equivariance in CNNs. Since images are usually square, these approaches often encode only discrete symmetry groups, such as four-fold rotational symmetries, sometimes including reflections \cite{laptev2016ti, cohen2016group, cohen2016steerable, weiler2019general, romero2019co, romero2020attentive, lafarge2021roto}. This is already a large improvement over the common method of rotating by various angles to augment the training data. Data augmentation leads to longer training times and gives no guarantees that the ML algorithm will encode the desired symmetry. More recent works have encoded continuous rotational symmetries by (projecting and then) working on the sphere \cite{cohen2018spherical, kondor2018clebsch, esteves2018learning,cohen2019gauge,defferrard2020deepsphere,esteves:2020:spin}.

Past works in wavelet scattering transforms have defined roto-translational invariant scattering networks and applied them to image classification problems\cite{sifre:2013:rotation}.\footnote{Previous works have also addressed scale-equivariant CNNs \cite{worrall2019deep,sosnovik2019scale} and wavelet networks \cite{sifre:2013:rotation, romero2020wavelet}, which we comment on only briefly.} However, we find that how "effectively invariant" the description provided by the network depends strongly on the design of the wavelets used and that it is advantageous for the Fourier transforms of the wavelets to sum to 1 in the Fourier domain. This advantage is preserved if the wavelet transform to the $p^{th}$ power sums to 1 as long as the transform outputs the modulus to $p^{th}$ power, summed. Most previous work chooses $p=1$. In the following, we choose $p=2$ in order to exploit the Parseval-Plancherel identity, and find good performance. In this work, we provide a fast-to-compute, translationally invariant and rotationally equivariant wavelet scattering network.\footnote{\url{https://github.com/andrew-saydjari/EqWS.jl}, see Sec. \ref{sec:dataavil}.} We then construct rotationally invariant statistics from the scattering network and quantify the invariance of those statistics. 


\section{Wavelet Design} \label{sec:WaveletDesign}

We introduce a new set of wavelets, triglets, which optimize the rotational equivariance of the scattering network. The full set of wavelets used in the scattering network are shown in Figure \ref{fig:FinkletFilterBank}. Other common complex wavelets are plotted side-by-side with these wavelets in Appendix \ref{sec:MorletOptim} for comparison. The triglets are defined in Fourier space by a cosine-windowing function on polar angle and log radius.

\begin{figure}[t!]
\centering
\includegraphics[width=0.98\linewidth]{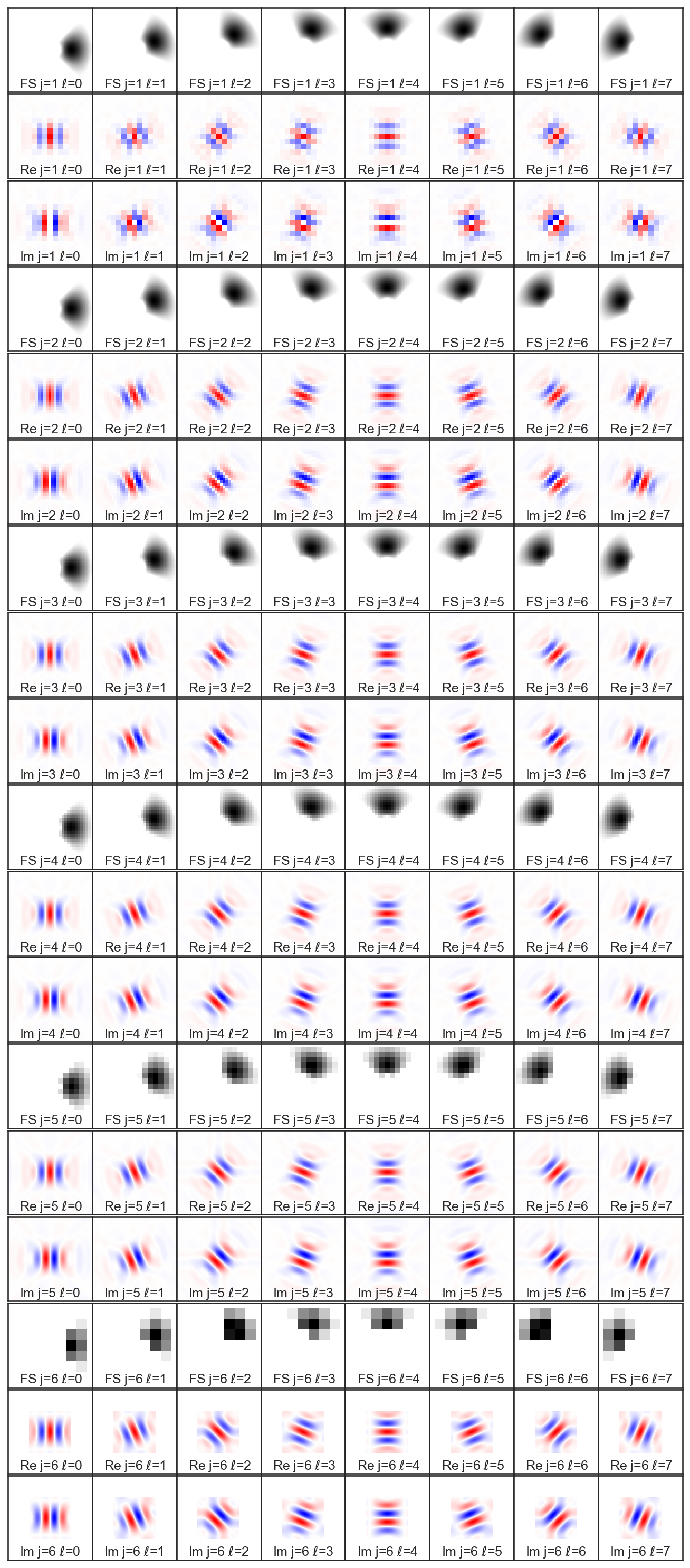}
\caption{Full filter bank of triglets for $L$ = 8, $c$ = 1, $w$ = 2, $p$ = 1, $N_{\rm{pix}}^2 = 256^2$. Note that unlike usual wavelet scattering networks, the total number of $j$ is not specified by choice of $J$, but by $c$ and the image size to be $c*(J^{\rm{im}}-2)$ where $J^{\rm{im}} = \log_2(N_{\rm{pix}})$ for an $N_{\rm{pix}}^2$ image. The Fourier space (``FS''), real part of the image-space (``Re''), imaginary part of the image space (``Im'') wavelets are shown indexed by $j$ and $\ell$. Fourier space wavelets are plotted with a scale of $2^{J^{\rm{im}-j}}$ and image space wavelets are plotted with a scale of $2^{j+3}$. Note for $j=6$ this is one logarithmic scale beyond the image size to remain consistent and emphasize edge effects. Fourier space plots are min-max normalized; dark colors represent high values. Real-space plots are symmetrically normalized around zero, with red positive and blue negative. }
\label{fig:FinkletFilterBank}
\end{figure}
\begin{figure}[t!]
\centering
\includegraphics[width=\linewidth]{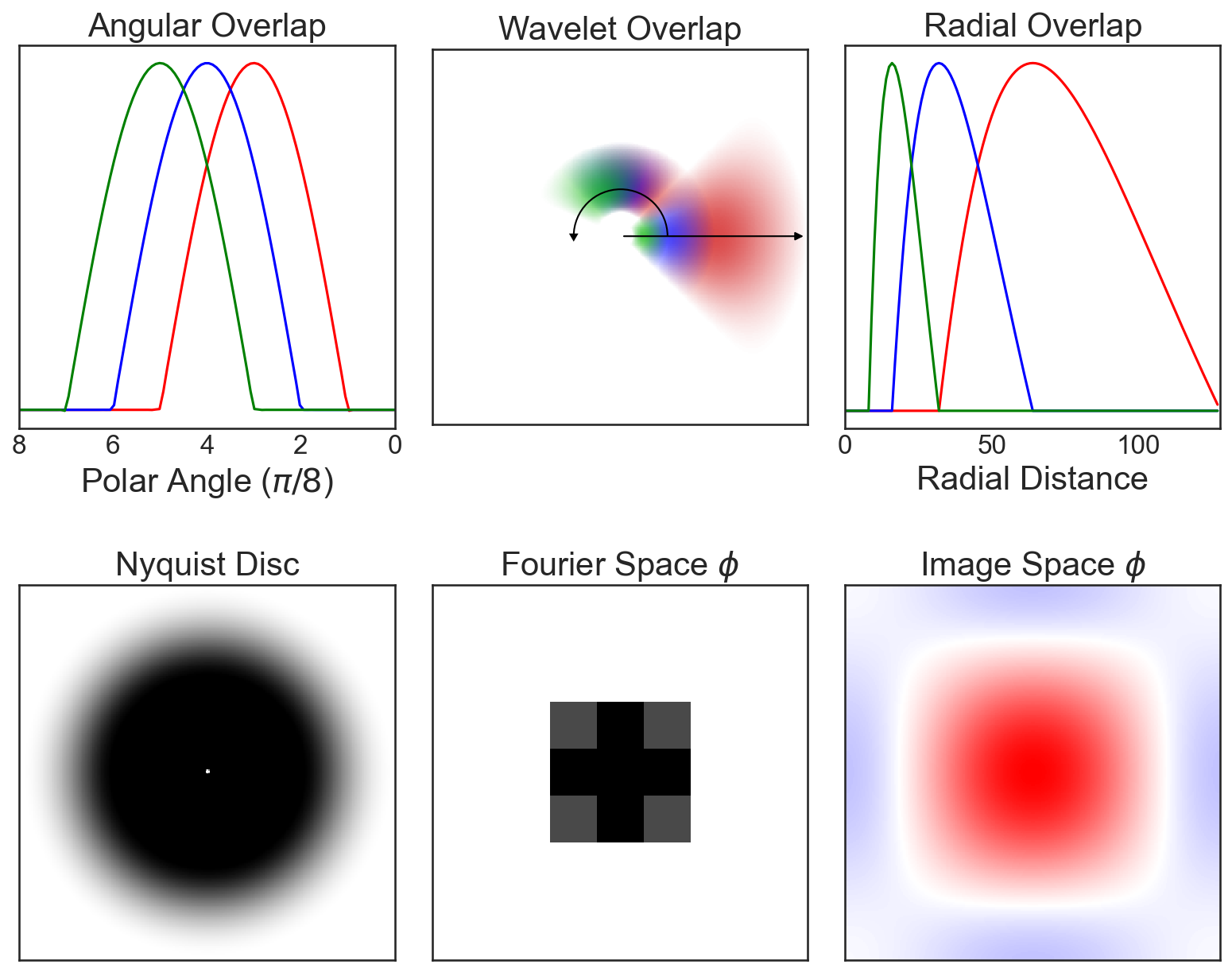}
\caption{Selected wavelets for $L = 8$, $c = 1$, $w = 2$, $t = 1$, $N_{\rm{pix}}^2 = 256^2$. Selected wavelets $j=1, 2, 3$ with $\ell=0$ in red, blue, and green respectively are shown in the \emph{top middle} in Fourier space to demonstrate the overlap of wavelets at different scales. A line cut along the radial direction at zero inclination is shown \emph{top right} with corresponding colors. Selected wavelets $j=2$ with $\ell=2, 3, 4$ in red, blue, and green respectively are also shown in the top middle panel in Fourier space. A line cut along the angular direction at $\rm{r}=32$ is shown \emph{top left} with corresponding colors. The sum of the squares of all $|\psi_{j,\ell}|^2$ are plotted in Fourier space \emph{bottom left}. The missing hole at the origin is filled by the $\phi$ filter which is shown in Fourier space and image space in the \emph{bottom middle} and \emph{bottom right} respectively. In Fourier space, $\phi$ is plotted with a scale of $2^{3}$ and in image space wavelets with a scale of $2^{8}$. Note that $\phi$ is not completely zero at the real space edge, partially breaking continuous rotational symmetry.}
\label{fig:FinkletOrient}
\end{figure}

\begin{ceqn}
\begin{align}\label{eq:finklet}
\widehat{\psi_{j,\ell}}(r,\theta) = 
\frac{1}{\sqrt{w}}
& \cos\left(
90\degree\left(
\log_2\left(r\right)-\left(J^{\rm{im}}-j-1\right)
\right)\right) \nonumber\\
& \cos\left(
\frac{L}{2wt}
\left(\theta-\ell t\frac{180\degree}{L}\right)
\right)\mathds{1}_\mathds{D}(r,\theta)
\end{align}
\end{ceqn}
Here $\mathds{1}_\mathds{D}(r,\theta)$ is the indicator function, which is $1$ everywhere in the domain where the arguments of both cosines are both $[-90\degree,90\degree]$. We define $J^{\rm{im}} = \log_2(N_{\rm{pix}})$ where $N_{\rm{pix}}$ is the size of one dimension of the image and we always assume images are square. As usual for wavelet scattering transforms, $j$ and $\ell$ label the logarithmic spatial scale and angular direction captured by the wavelet. Like previous definitions of wavelet scattering networks, $\ell \in [0,L-1]$. However, we restrict $j \in [1,1+1/c, ... , (J^{\rm{im}}-2)]$. The $j=0$ wavelets are not well sampled in image space and unevenly sample directions (i.e., they extend into the corners) because they peak at $r=2^{J^{\rm{im}}-1}$ in Fourier space. Thus, we exclude the $j=0$ wavelets entirely. Similarly, the $J^{\rm{im}}-1$ wavelets are not well-sampled in Fourier space. We replace them with one wavelet, $\phi$, centered at the origin of Fourier space to capture all remaining power in the Nyquist disc. All power at wavenumber zero is captured by $\phi$, i.e., it is the only filter that has non-zero mean. It is rotationally symmetric up to pixelation effects, and is real in image space (Figure \ref{fig:FinkletOrient}). 

\begin{table}[hb]
\centering
\caption{Variable Reference}
\begin{tabular}{clc}
\toprule
Parameter & Description & Default Value\\

\midrule
$c$ & Radial bin spacing $2^{1/c}$ & $1$ \\
$j$ & Radial bin index & \\
$J^{\rm{im}}$ & $\log_2$ of image size ($2^{J^{\rm{im}}} \times 2^{J^{\rm{im}}}$ pixels) & \\
\midrule
$L$ & Number of angular bins & $8$ \\
$\ell$ & Angular bin index &  \\
$w$ & Angular width in multiples of $180\degree/L$ & $2$ \\
$t$ & Subdivides Fourier half($1$)/full($2$) plane & $1$ \\
$m$ & Depth of scattering network & $2$ \\
\midrule
$r$ & Fourier-space radial coordinate & \\
$\theta$ & Fourier-space angular coordinate & \\
$\vec x$ & Real-space coordinates & \\

\bottomrule
\end{tabular}
\end{table}

The parameter $c$ determines how tightly spaced the wavelet centers are in logarithmic radius, where they all extend $\pm 1$ logarithmic radial bin in Fourier space (Figure \ref{fig:FinkletOrient}). For $c=1$, the usual dyadic spacing is obtained, while for larger integer $c$, there are $c \times (J^{\rm{im}}-3) + 1$ values of $j$.

The parameter $w$ determines the angular width of the wavelets in multiples of $180\degree/L$. In order to compare to infinitely wide CNNs or to obtain high angular precision, we would like to be able to take the large $L$ limit. However, large $j$ wavelets near the Fourier space origin easily become poorly sampled in the angular direction in this limit. To mitigate this problem, $w$ is dynamically adjusted as a function of $j$ to the smallest possible integer so that Eq. \ref{eq:widthCond} is satisfied. There is a qualitative trade-off between the sharpness of the angular response and requiring the wavelets to be well sampled, which must be made in choosing a scalar prefactor for the right-hand side of Eq. \ref{eq:widthCond}.
\begin{ceqn}
\begin{align} \label{eq:widthCond}
w > \frac{L}{2^{J^{\rm{im}}-j-1} \times t \times 180\degree}
\end{align}
\end{ceqn}
The parameter $t$ determines whether or not the total of $L$ angular divisions subdivides the Fourier half plane ($t=1$) or full plane ($t=2$). Because the Fourier transform of a real image is conjugate symmetric, $F(x) = F^{*}(-x)$, where $F^{*}$ is the complex conjugate of $F$, wavelet coefficients in the lower-half plane are completely redundant as a result of the modulus between layers of the scattering network.

Triglets are closely related to ridgelets, steerable wavelets, and curvelets, which all differ from each other in the definition of angular divisions and use a piece-wise cosine windowing function \cite{ma2010curvelet}. The triglets, based solely on cosines, go to zero at wavenumber zero (except for $\phi$), and therefore are mean zero in the real domain. They are easily modified to study pooling of the image pixels to arbitrary powers $p$, though we only study $p=2$ here. Triglets do not have a flat top, a region where the window has constant response. This eliminates cases where changes in the test image scale or angular orientation lead to no change in the wavelet coefficients (because the change in wavevector only redistributes power within the flat region). We view this as a beneficial change for the application here since such cases likely make learning scale and angular information from the coefficients with machine learning techniques more difficult.

The key motivation for the design of these wavelets is that the sum of the squares $|\psi_{j,\ell}|^2$ is uniform in the Fourier domain (Figure \ref{fig:FinkletOrient}). A slight modification of the definition of the wavelet scattering coefficients, which we call equivariant wavelet scattering (EqWS), is required to make optimal use of this fact. The usual scattering coefficients are defined as 
\begin{ceqn}
\begin{align}
S_0^{\rm{orig}} & =  \int I(\vec x) \;d^2\vec x \qquad\\ \nonumber
S_1^{\rm{orig}}(j_1,\ell_1) & =  \int |I \star \psi_{j_1,\ell_1}|(\vec x) \;d^2\vec x \\ \nonumber
S_2^{\rm{orig}}(j_1,\ell_1,j_2,\ell_2) & = \int ||I \star \psi_{j_1,\ell_1}| \star \psi_{j_2,\ell_2}|(\vec x) \;d^2\vec x \nonumber
\end{align}
\end{ceqn}
We modify this definition\footnote{The wavelet scattering transform definition sometimes appears with an additional normalization factor which fixes the response with respect to real-space delta functions \cite{Bruna:2012:arXiv:}; the normalization is often dropped in practice due to subsequent transformations (such as taking the logarithm) of the scattering coefficients \cite{Allys:2019:A&A:}. Since we require a stronger point-wise normalization condition, that the sum of $|\psi_{j,\ell}|^p$ is uniform in the Fourier domain, we drop these normalization factors.} for EqWS in order to focus on power in the Fourier domain and conservation of total image power at each layer.\footnote{In reference to conservation, image "power" is often referred to as image "energy."} While the scattering network is still a tower of convolutions composed with the modulus function, we output the sum of the image power after those operations as the coefficient at each layer.
\begin{ceqn}
\begin{align}
\mu_0 & =  \int I(\vec x) \;d^2\vec x \qquad\\ \nonumber
\sigma^2_0 & = \int |I(\vec x)-\mu_0|^2 \;d^2\vec x \qquad\\ \nonumber
S_1(j_1,\ell_1) & =  \int |\bar I \star \psi_{j_1,\ell_1}|^2(\vec x) \;d^2\vec x \\ \nonumber
S_2(j_1,\ell_1,j_2,\ell_2) & = \int ||\bar I \star \psi_{j_1,\ell_1}| \star \psi_{j_2,\ell_2}|^2(\vec x) \;d^2\vec x \nonumber
\end{align}
\end{ceqn}
Here $\bar I(\vec x)$ is the mean-zero, unit-variance normalized image. Note that the $S_0$ term includes both the image mean and variance, so $S_0$ is two dimensional. After this magnitude information is recorded, the image is normalized prior to all convolutions. This should improve the comparison of coefficients between different images and strip out magnitude information when a focus on structure is desired.\footnote{We are especially interested in this case when trying to compare simulations and observations where the structure is often correct, but the magnitude may be in different units.}

Transformations on the image which preserve the image power also conserve the sum of all the scattering coefficients by the Parseval-Plancherel identity. This follows directly from the definition for $S_1$ and holds for $S_2$ because the modulus is norm-preserving; the logic for $S_1$ then goes through using $|\bar I \star \psi_{j_1,\ell_1}|$ as the image field. As an added benefit, $S_1$ can now be interpreted as just logarithmically spaced and angularly separated bins of the Fourier power spectrum. Since the sum of the squares of the wavelets is flat in Fourier space and the power at a given radius is partitioned equally between the wavelets with different $\ell$, EqWS has exact equivariance with respect to rotations that are multiples of $180\degree/L$. A rotation of an image by $s \times 180\degree/L$ simply permutes all angular indices of a given scattering coefficient.
\begin{align}\label{eq:equivPermute}
S_x(j_1,\ldots,\ell_1,\ldots) = S_x(j_1,\ldots,\ell_1+s,\ldots)
\end{align}
This exact discrete symmetry can better approximate a continuous rotational symmetry in the limit of large L. However, we also will demonstrate to what extent EqWS is approximately rotationally equivariant for angles which are not multiples of $180\degree/L$ in Section \ref{sec:RotEq}.

With this equivariance, we can also construct rotationally invariant coefficients (``ISO'') and guarantee that $S_1^{\rm{iso}}$ is as invariant to image rotations as the image power is. Analogously, $S_2^{\rm{iso}}$ can be defined by summing over one angular index.
\begin{ceqn}
\begin{align}\label{eq:isodef}
S_1^{\rm{iso}}(j_1) & =  \sum_{\ell_1=0}^{L-1} S_1(j_1,\ell_1) \\ \nonumber
S_2^{\rm{iso}}(j_1,j_2,\Delta_\ell) & = \sum_{\ell_1=0}^{L-1} 
S_2\left(j_1,\ell_1,j_2,\ell_1+\Delta_\ell\right) \nonumber
\end{align}
\end{ceqn}
Where $\Delta_l = \ell_2-\ell_1 \in [0,7]$ and all angular arguments are taken mod $L$. 

The $S_2^{\rm{iso}}$ coefficients are not exactly invariant under rotation, even in the case of exactly rotationally invariant wavelets. We can think of the ``true'' $S_2^{\rm{iso}}$ coefficient at some $(j_1,j_2)$ as an integral through $(\ell_1,\ell_2)$ space at fixed $\Delta_\ell$, which the sum approximates. To the extent that the $S_2$ function of $(\ell_1,\ell_2)$ is well sampled, the sum is a reasonable approximation of the desired integral. By taking $w=2$ or higher in Eq \ref{eq:finklet}, we smooth the angular response of the wavelets, effectively broadening features in the $(\ell_1,\ell_2)$ plane. 

In order to make these claims of rotational equivariance, two assumptions were made:
\begin{enumerate}[label=(\roman*)]
\item In Fourier space, no power is outside the Nyquist disc, the circle of radius $N_{\rm{pix}}/2$.
\item In image space, all pixels outside the circle of radius $N_{\rm{pix}}/2$ are zero.
\end{enumerate}

If there is power outside the Nyquist disc and that power is simply in the corners of the Fourier plane (i.e. not aliased), we can achieve the first condition by resampling the image at a finer resolution. Aliased power is more insidious because resampling with any reasonable interpolation algorithm will not be able to recover that power and we must simply admit that the method is rotationally invariant up to the resolution of the data. For the second condition, we will compromise and use an apodization filter to set the corners to zero if they are not already.\footnote{We implement a Tukey filter with $\alpha=0.3$, which means that $70\%$ of the radial extent of the filter is flat before it falls to zero as a cosine.} This can be problematic for image classification tasks on small images cropped closely to the object of interest. However, in most cases we can simply change the subdivision of a larger field of view so that the circularly cropped image after the apodization filter contains the objects or field of interest. This limitation replaces interpolating images onto a sphere, which is common in the equivariant CNN literature.\footnote{Note that here we have only one free parameter, $\alpha$ of the apodization function, which replaces the free parameters associated with that spherical projection.}

The modifications introduced above in defining EqWS allow us to achieve significant speed-ups ($\sim$40x faster per coefficient) relative to currently public scattering network packages such as \textsc{Kymatio}.\footnote{\textsc{kymatio} \cite{Andreux:2018:arXiv:} is a \textsc{Python} implementation of the WST available at \url{https://www.kymat.io}} Detailed timings providing comparisons to the code released with this paper, EqWS.jl, are shown in Appendix \ref{sec:CompCost}. By outputting the pooled power instead of absolute value of the convolved fields, the Parseval-Plancherel identity permits pooling in Fourier space instead of real space, saving order $(J \times L)^m$ inverse Fourier transforms for a scattering network with $m$-layers. By using compact wavelets in Fourier space, the convolutions can be computed as sparse matrix multiplications and benefit from reduced memory allocation. However, by choosing to pool images into coefficients under the modulus to the $p$ power where $p=2$ instead of $p=1$, EqWS up-weights the contribution of large pixel values. One generic downside is that this allows outliers to cause larger perturbations. However, the optimal choice of $p$ is likely application dependent.


\section{Explicit Equivariance Tests} \label{sec:EquivTests}

Since the Fourier transform is translation equivariant, phase-independent convolution methods sometimes implicitly assume translation invariance. However, as mentioned above, subsampling---such as strided-convolution and pooling--- common practices in the machine learning community can break this invariance. We explicitly checked that EqWS is translation invariant by initializing a random $128 \times 128$ pixel image with entries in the interval $[0, 1)$. This matrix was then shifted with periodic boundary conditions by every integer number of pixels. The maximum change of any scattering coefficient relative to those of the original image was $2 \times 10^{-16}$, which closely matches the numerical precision of the FFT and validates our translation invariance.

\begin{figure}[t!]
\centering
\includegraphics[width=0.965\linewidth]{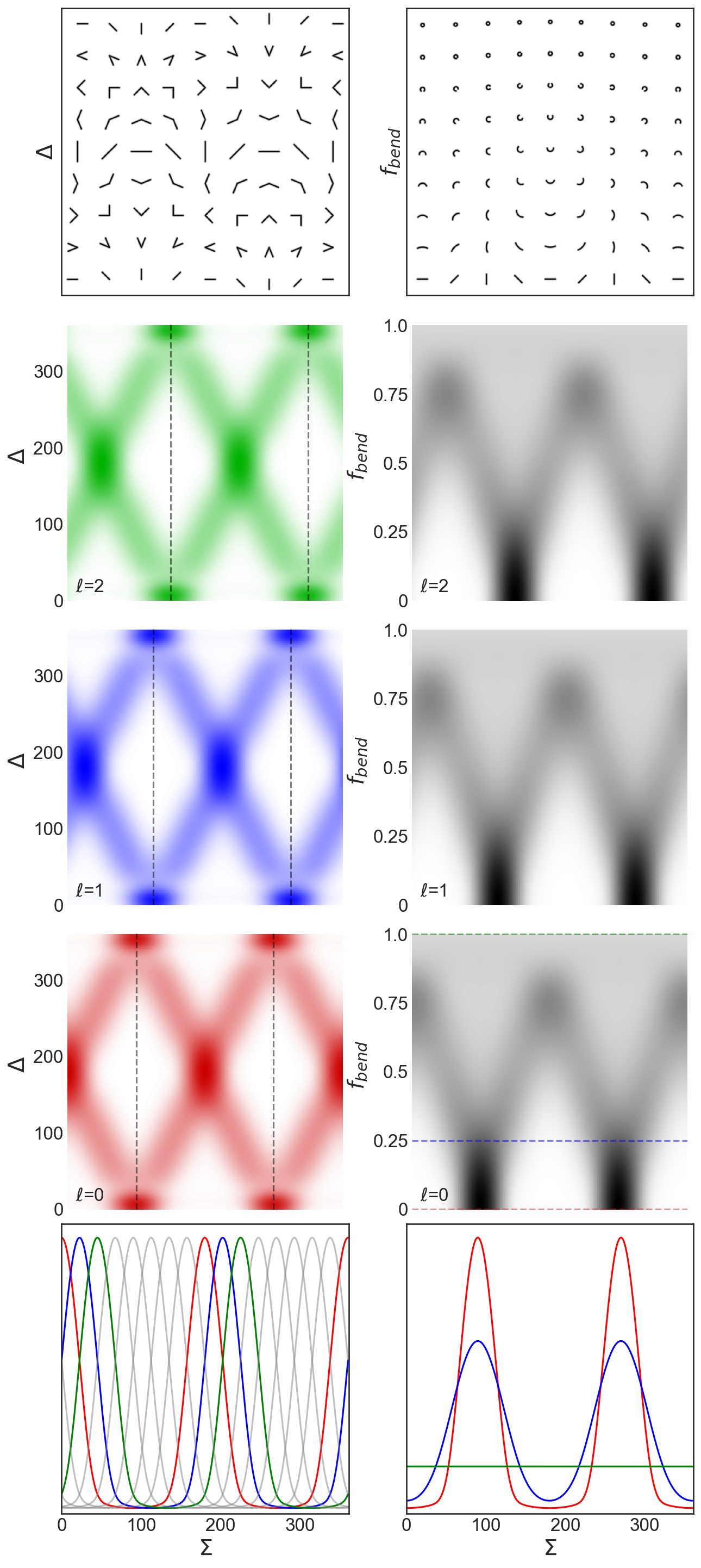}
\caption{Demonstrating rotational equivariance on test images. \emph{Top left:} Matrix that subsamples the test rod images testing corners at different angles. \emph{Middle left:} Maps of the magnitude of $S_1(j=4,\ell))$ for $\ell=0, 1, 2$ for images of the test rods above; darker colors represent larger values. Dashed vertical lines indicate expected peak at $\Delta = 0$ for a given $\ell$. \emph{Bottom left:} Line cuts along $\Delta = 180\degree$ for every $\ell$. Colored profiles correspond to the same colored plot above, plotted on the same y-scale starting from zero. \emph{Top right:} Matrix that subsamples the test rod images testing circular deformation at different angles. \emph{Middle right:} Maps of the magnitude of $S_1(j=4,\ell)$ for $\ell=0, 1, 2$ for images of the test rods above; darker colors represent larger values. \emph{Bottom right:} Line cuts for $\ell=0$ at $f_{\rm{bend}}=0, 0.25, 1.0$ as indicated by the dashed lines on the plot above, plotted on the same y-scale starting from zero. Both tests show smooth, equivariant, angular response of $S_1(j=4,\ell))$.
}
\label{fig:RotEquiv}
\end{figure}

\begin{figure}[b!]
\centering
\includegraphics[width=0.98\linewidth]{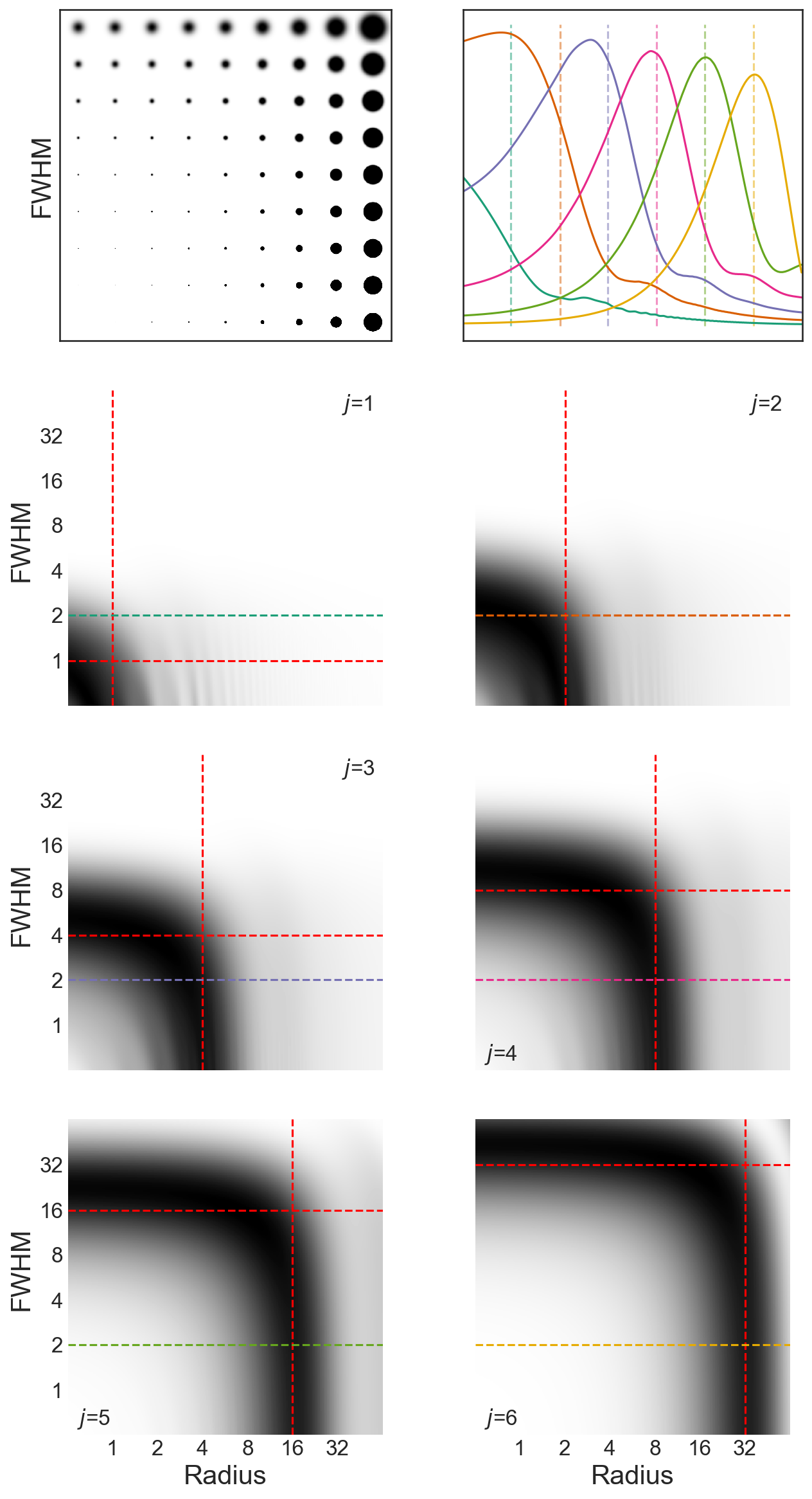}
\caption{Demonstrating scale dependence on test disc images. \emph{Top left:} Matrix that subsamples test disc images, varying radii and FWHM. \emph{Bottom:} Maps of the magnitude of $S_1(j,\ell=0)$ for images of test discs above; darker colors represent larger values. Dashed red lines indicate expected peak at response at $2^{j-1}$. \emph{Top right:} Line cuts along FWHM $ = 2$ for every $j$. Colored profiles correspond to colors of dashed lines on the lower plots and represent $j = 1$ to $6$ left to right. Vertical dashed lines mark expected peak at $2^{j-1}$. These tests show $j$ can be easily interpreted as a scale index and suggest that approximate equivariance could be achieved by excluding the largest and smallest $j$ scales.
}
\label{fig:ScaleEquiv}
\end{figure}

In order to better understand the equivariance of EqWS, we demonstrate the response of the scattering coefficients to simple test images. Because the triglets and subsequent scattering coefficients are labeled by angular direction and spatial scale, the tests are designed to probe how interpretable those scales and angles are with respect to features in an image. Interactive versions of Figures \ref{fig:RotEquiv} and \ref{fig:ScaleEquiv} showing the results for arbitrary coefficients is available online for \href{https://faun.rc.fas.harvard.edu/saydjari/EqWS/interactive_corners.html}{\textbf{test corners}}, \href{https://faun.rc.fas.harvard.edu/saydjari/EqWS/interactive_curves.html}{\textbf{test curves}}, and \href{https://faun.rc.fas.harvard.edu/saydjari/EqWS/interactive_discs.html}{\textbf{test discs}}.\footnote{\url{https://faun.rc.fas.harvard.edu/saydjari/EqWS/}}

First we consider the angular equivariance using images of rods. Each test image is $256 \times 256$ pixels and contains two rod segments. Each rod segment has one end fixed at the origin, is $40$ pixels long, and is broadened by a Gaussian envelope (FWHM = 6).\footnote{The full width at half maximum (FWHM) is a measure of the width, equal to $2.355 \sigma$, which we report in pixel units.} The rods are at angles $\Sigma+\Delta/2$ and $\Sigma-\Delta/2$ so that they have an opening angle $\Delta$ and net orientation $\Sigma$. A subsample of the test rods are placed in a matrix in Figure \ref{fig:RotEquiv} to help visualize the test image space. Figure \ref{fig:RotEquiv} shows the amplitude of $S_1(j=4,\ell)$ in response to the test rods for $\ell=0, 1, 2$, $\Sigma \in [0,360\degree)$, and $\Delta \in [0,360\degree)$. 

The response of each $S_1(j=4,\ell)$ are offset with respect to $S_1(j=4,\ell-1)$ by $180\degree/L$ as expected. For $\Delta = 180\degree$, we reduce to the single rod case and note that the vertical rod results in the largest response in the $\ell=0$ coefficient. This follows from the fact that the wavevector associated with a rod is perpendicular to it and, in this case, is thus along the $x$-axis. In addition to the $\ell=0$ coefficient which is largest, $\ell=1$ has nonzero magnitude while $\ell=2$ is zero. This illustrates the fairly sharp and sparse angular response of the triglets. This response is controlled both by $L$, the total number of angular divisions, and $w$, the angular width of the wavelets in multiples of $180\degree/L$. Here we chose $w=2$ to enhance smooth variations of the coefficients under rotation and $L=8$ for convenience, but a larger choice of $L$ would give sharper angular response. 

This peak in the $\ell=0$ coefficient is replicated at both $\Sigma = 0$ and $180\degree$. While the rod is symmetric about the origin and thus invariant under net rotations of $180\degree$, this symmetry under $180\degree$ rotation extends beyond the case of highly symmetric objects. Because we are working with real images and the scattering network pools under the modulus, the scattering coefficients for images which are $180\degree$ rotations of one another are identical, as their Fourier transforms differ only by a global $180\degree$ phase. As $|\Delta-180\degree|$ increases, the peak response of the rod splits for a given $\ell$ to the expected $\Sigma$ for peak response to each arm. At $\Delta = 0$ and $360\degree$ we recover the response at $\Delta = 180\degree$ with a $90\degree$ shift, since a single rod is formed from the superposition of the two rods. Dashed vertical lines are superimposed on the three maps of $(\Sigma,\Delta)$ space for $\ell=0, 1, 2$ to indicate the expected peak response at $\Delta = 0$ for a given $\ell$, $180\degree \times \ell/L + 90\degree$. For some of the small-scale ($j$) coefficients (in interactive figure, not shown here), there is a small dip in magnitude for the coefficients just before the two arms of the rod are fully superimposed. This can be attributed to the effectively wider, single rod that resonates more with a different $j$-scale. To emphasize the equivariance, we show line cuts at $\Delta = 180\degree$ in the bottom panel for all of the $\ell \in [0,7]$ to demonstrate that all angular orientations are equivalent. 

A comparison with Figure 21 from \cite{selesnick:2005:dual} shows that high-quality wavelet transform procedures, which are well-optimized for different goals, often employ wavelets which still reflect the rectangular grid and can only be equivariant with respect to $90\degree$ rotations.\footnote{For many applications, $90\degree$ symmetry may be all that is practical or required. Images on a rectangular grid will always have some continuous rotational symmetry breaking, this work shows how wavelet design can mitigate this breaking of symmetry.} A comparison to \cite{cammarata:2020:curve} shows that EqWS has more rotationally equivariant response as compared to layer 3b in InceptionV1, a CNN trained on ImageNet. However, the activation of these CNN layers successfully breaks the $180\degree$ rotational symmetry inherent in EqWS.

We are also interested in how the angular behavior of EqWS coefficients continuously evolves as the image being analyzed becomes itself rotationally invariant. To probe this evolution from equivariance to invariance, we study another set of test images which deform a single rod (length$ = 20$, FWHM$ = 6$) to a ring. The deformation leaves the midpoint of the rod fixed at the origin and is parameterized by $f_{\rm{bend}} = 1/r_{\rm{curvature}}$ with $f_{\rm{bend}} \in (0,1]$ and $\Sigma$, the direction of the center of curvature. A subsample of the test images is placed in a matrix in Figure \ref{fig:RotEquiv}, top right. Figure \ref{fig:RotEquiv}, right shows the amplitude of $S_1(j=4,\ell)$ in response to the test arcs for $\ell=0, 1, 2$, $\Sigma \in [0,360\degree)$, and $f_{\rm{bend}} \in [0,1)$. 

As $f_{\rm{bend}}$ increases from zero (which reproduces the $\Delta = 0$ cut on the left panels), the peak in $S_1(j=4,\ell)$ as a function of $\Sigma$ at $180\degree \times \ell/L + 90\degree$ broadens until $f_{\rm{bend}} = 0.5$. This corresponds to the power from the rod being distributed over a wider range of wavevectors. Beyond $f_{\rm{bend}} = 0.5$, the peak in $S_1(j=4,\ell)$ occurs at a $180\degree$ shift from the peak at $f_{\rm{bend}} = 0$ which likely is due to Fourier power from the gap in the circle, which is locally an empty bar surrounded by saturated pixels and is perpendicular to the original rod direction. The detailed behavior for $f_{\rm{bend}} \in (0.5,1.0)$, especially for small $j$, depends on geometric parameters (FWHM, length) and the interpolation parameterization so we do not analyze it further. 

In the limit of $f_{\rm{bend}}=1$, the rod becomes a ring which is rotationally invariant. A rotationally invariant image should equally activate all wavelets that differ only by a rotationally equivariant index. To demonstrate this limit, we take line cuts for $\ell=0$ at $f_{\rm{bend}}=0, 0.25, 1.0$ as indicated by the dashed lines on the $(\Sigma,f_{\rm{bend}})$ plot (Figure \ref{fig:RotEquiv}, bottom right). The cut corresponding to $f_{\rm{bend}}= 1.0$ is flat as expected, demonstrating that there is no angle dependence in the response of EqWS to the circle. In attempting to compare how rotationally equivariant different methods are, it is important to benchmark how flat this response is. We find that the fractional fluctuations (standard deviation over the mean) in the $S_1(j=4,\ell)$ are on the order of $2\times10^{-3}$ while the fluctuations in image power due to specifying the function on a grid are $3\times10^{-4}$. We believe coefficient fluctuations only an order of magnitude worse than the fluctuations resulting from pixelating the function should be sufficient for most applications. A more detailed display of the second-order ISO EqWS coefficient stability (for all $j$, $\ell$) is in Appendix \ref{sec:CoeffStab}.

We investigate the scale-dependent response of EqWS using $256 \times 256$ pixel images of discs. Each disc is centered at the origin and specified by a radius ($r$) and a Gaussian broadening scale (FWHM). We logarithmically sample these two scale axes ($[0.5, 64]$) because the wavelets logarithmically sample the radial coordinate in Fourier space. Some representative test discs are placed in a grid in Figure \ref{fig:ScaleEquiv} (top left) to help visualize the test image space. Figure \ref{fig:ScaleEquiv} (bottom) shows the amplitude of $S_1(j,\ell=0)$ in response to the test disc for all $j$ on $\log_2$ axes. Red dashed lines serve as a guide to the eye at $2^{j-1}$ on both axes. Colored dashed lines correspond to line cuts taken at FWHM$= 2$ show in Figure \ref{fig:ScaleEquiv} (top right). 

For $j=1$ and $j=2$, the Radius-FWHM plot shows significant ringing at low FWHM. This is indicative of the image not being well sampled. These abrupt changes in value for adjacent pixels cause oscillations in Fourier space that predominantly affect the largest Fourier scales, and thus smallest $j$. For a disc, these are associated with the well known oscillations of Bessel functions. For all $j$ we observe a corner-like feature in the Radius-FWHM plot, which is approximately centered at $2^{j-1}$ on both axes. The continuous Fourier transform of a disc of radius $R$ in a $256 \times 256$ pixel image is proportional to $J_1(2 \pi r/256)/r$, where $r$ is the Fourier space radial component and $J_1$ is the Bessel function of the first kind. If we suppose the peak response of the $S_1(j,\ell)$ coefficient will occur when the first peak of $J_1(2 \pi r/256)/r$ coincides with the peak of the wavelet in Fourier space, we predict a peak response at a disc radius of $\sim 2^{j-1}$.

Deviations from the above rule occur upon relaxing the assumptions we made. For example, the magnitude of the $S_1$ coefficients is actually an integral of the Fourier image power and wavelet amplitude of over the wavelet support, and the peak of the integral need not correspond to when a local maximum of the Fourier image power and wavelet peak are at the same point, as assumed in this heuristic discussion. Using the exact roots of the derivative of $J_1(2 \pi r/256)/r$ predicts a peak response at radii slightly larger than $\sim 2^{j-1}$, especially for low $j$ (small radii). Deviations of the discrete Fourier transform of a pixelated disc from $J_1(2 \pi r/256)/r$ similarly predict a peak response at radii slightly larger than $\sim 2^{j-1}$. However, a pixelated disc is a poor approximation of a disc, or of real images which are smoothed by some point-spread function (PSF) when well sampled. This pixelation also introduces oscillations on top of the $S_1$ response envelope as discussed below. To make the disc images well sampled, we blur each image with a Gaussian PSF, by multiplication of the Fourier domain by a Gaussian envelope of inverse width. The PSF suppresses power at large Fourier radii, shifting the peak in the Fourier power (not at the origin) to smaller Fourier radii (larger $j$). For fixed radius, we see that $S_1(j,\ell)$ respond most strongly to a disc when it is broadened by a Gaussian PSF with $2^{j-1}$ FWHM.

We choose to compare line cuts at FWHM$= 2$ so that the images are well sampled. For those line cuts, the amplitude of $S_1(j,\ell=0)$ peaks approximately at $2^{j-1}$ for $j=4, 5, 6$, where the $2^{j-1}$ values are marked with vertical dashed lines. Shoulders at approximately $2^{j+1}$ are observed, resulting from the higher-order peaks of the Bessel function. At smaller $j$, the peak of the line cuts precedes $2^{j-1}$ as a result of the FWHM being comparable to $2^{j-1}$ (above or near the corner in Radius-FWHM space), blurring the image and suppressing power which overlaps with those $j$-scale wavelets. Similar line cuts at FWHM$= 0.5$ show peaks at approximately $2^{j-1}$ for all $j$ except $j=1$, but at the cost of oscillations in amplitude for even more of the low $j$ coefficients. 

While we leave achieving a notion of scale equivariance to later work, these results show that the scales indexed by $j$ are highly interpretable. However, attempts at scale equivariance or invariance will likely need to restrict to well-sampled images in real space to damp ringing in Fourier space and have smoothly varying EqWS coefficients as a function of scale. Thus the first few $j$ will likely have to be excluded. To optimize being well sampled in Fourier space, an apodization function, such as a Gaussian, may be optimal since multiplication in real space by a Gaussian is simply Fourier space convolution and will thus smooth the Fourier space image.\footnote{This in part gives up exact translation invariance even within the center of the apodization window because a Gaussian is nowhere constant. This trade-off between symmetries will need to be evaluated for a given application.} Of course, the broad response of the coefficients shown in Figure \ref{fig:ScaleEquiv} (top right) illustrates that both the largest and smallest scales must be excluded in order to well approximate scale invariance.\footnote{While some attempts to sum over $j$ and achieve scale invariance have been made (see Eq. 20 in Ref. \cite{sifre:2013:rotation}), these likely suffer from the effects in Figure \ref{fig:ScaleEquiv} (top right).} \\


\section{Linear Learning} \label{sec:LinLearn}

\begin{table*}[ht!]
\centering
\caption{MNIST Accuracy of EqWS+LDA with Different Train/Test Angles}
\begin{tabular}{c|cccc|cccc}
\toprule

\multirow{2}{*}{Train Angle} & \multicolumn{4}{c|}{Test Angle- REG} & \multicolumn{4}{c}{Test Angle- ISO} \\
& $0$ & $7/8 \times 180\degree$ & $4/3 \times 180\degree$ & $n/3 \times 180\degree$ & $0$ & $7/8 \times 180\degree$ & $4/3 \times 180\degree$ & $n/3 \times 180\degree$ \\

\midrule
$0$ & $96.23$ & $83.30$ & $25.19$ & $48.16$ & $93.50$ & $88.37$ & $85.07$ & $87.67$ \\
$7/8 \times 180\degree$ & $82.16$ & $96.14$ & $18.96$ & $52.76$ & $86.88$ & $93.36$ & $92.87$ & $86.23$ \\
$4/3 \times 180\degree$ & $22.70$ & $15.83$ & $96.16$ & $48.12$ & $82.25$ & $92.81$ & $93.43$ & $86.12$ \\
$n/3 \times 180\degree$ & $94.12$ & $84.01$ & $94.16$ & $94.13$ & $91.95$ & $92.14$ & $92.25$ & $92.05$ \\
\bottomrule
\end{tabular}
\end{table*}

We benchmark EqWS, an equivariant wavelet scattering network, on standard machine learning datasets below using the simplest regression and classification techniques. In general, we will use linear regression for parameter estimation and linear discriminant analysis (LDA) for classification. Linear regression with a one-hot encoding performs nearly as well at classification as LDA, but we prefer LDA since it accounts for the distance of a point to all class means simultaneously. Undoubtedly, nonlinear classification schemes (e.g., a dense neural network) acting on the EqWS outputs have the capacity to outperform a linear classifier.  We do not consider more advanced regression and classification approaches in this work, but rather focus on the descriptive power of the nonlinearity in EqWS and the equivariance or invariance thereof.

\subsection{Rotational Invariance} \label{sec:RotInvar}

We first focus on MNIST and create rotated versions of it using bi-cubic interpolation. In order to work with well-sampled images and easily apply circular apodization without cutting off any portion of the digits, the standard MNIST images were both padded and interpolated. The digits ($28 \times 28$) were embedded in a $128 \times 128$ image and a bi-linear interpolation to a $256 \times 256$ image was then performed. See Appendix \ref{sec:CompCost} for the trade-off between computational time and rotational invariance that informed this choice. 

We compare the full equivariant set of coefficients (REG, 2452 dimensional) and the reduced rotationally invariant coefficients (ISO, 310 dimensional; see Eq \ref{eq:isodef}) when the train and test images are rotated to the same or different angles ($0, 7/8 \times 180\degree, 4/3 \times 180\degree)$. When the train and test images are at the same angle, LDA achieves an accuracy of $96\%$ on the REG coefficients and $93\%$ in the ISO coefficients. This slight decrease in the ``clean'' accuracy follows from the fact that the total orientation can be useful in digit classification, for example in distinguishing ``6'' and ``9.''\footnote{Even using a totally rotationally invariant set of coefficients, ``6'' and ``9'' can often still be distinguished in handwritten digits, for example by the straighter stem of the ``9.''} This is an instance of the general observation of decreased ``clean'' accuracy in robust classification algorithms.

For angles such as $0$ and $7/8 \times 180\degree$ which are similar under the $180\degree$ symmetry of EqWS, training on one and testing on the other only leads to a drop of about $13\%$ in accuracy for the REG coefficients. However, training but testing on angles that differ modulo $180\degree$ by a significant fraction of $180\degree/L$ ($4/3 \times 180\degree$ and $7/8 \times 180\degree$) results in an accuracy below $20\%$. In contrast, for the ISO coefficients, only a slight dependence on the distance between test-train angle is observed, and the minimum accuracy is $82\%$. This comparison already shows that classification with ISO coefficients is more robust to rotations, but is slightly underwhelming given that the triglets perfectly cover the Nyquist disc by design.

As a comparison to baseline methods of robust training, we train using data augmentation at all multiples of $180\degree/3$ (3 angles). For REG, we see a small decrease in the ``clean'' performance, performance on angles used in training, to $94\%$ and a drop of around $10\%$ on angles far from the train angles. For ISO, we see a small decrease in the ``clean'' performance to $92\%$, but see no clear change in performance on angles far from the train angles. This improvement for ISO training on six angles relative to training on one angle comes from the small remaining fluctuations of the coefficients (discussed below). Training at even one additional angle (which is a significant fraction of $180\degree/8$ away from the first training angle) recovers the expected invariance, which is a minimal and practical amount of data augmentation.

\begin{table}[hb]
\centering
\caption{Accuracy of EqWS+LDA on Benchmark Datasets} 
\begin{tabular}{l|lll}
\toprule
 & NR/NR & R/R & NR/R \\

\midrule
EqWS+LDA REG & $96.23$ & $92.3(1)$ & $50.2(5)$ \\
EqWS+LDA ISO & $93.5$ & $92.10(5)$ & $87.7(2)$ \\
EqWS+LDA REG (3 train angle) & $94.15$ & $92.3(1)$ & $88.4(2)$ \\
EqWS+LDA ISO (3 train angle) & \bm{$91.95$} & \bm{$92.10(5)$} & \bm{$92.12(8)$} \\
Cohen et al. \cite{cohen:2018:spherical} & $95.59$ & $94.62$ & $93.40$ \\
Kondor et al. \cite{kondor:2018:clebsch} & $96.4$ & $96.6$ & $96.0$ \\
Esteves et al. \cite{esteves:2020:spin} & $99.37(5)$ & $99.37(1)$ & $99.08(12)$ \\
Planar CNN \cite{esteves:2020:spin} & $99.07$ & $81.07(63)$ & $17.23(71)$ \\
\bottomrule
\end{tabular}
\label{table:NR_R}
\end{table}

We then perform the usual test of training on one orientation and testing on random rotations (Table \ref{table:NR_R}). We provide for comparison state-of-the-art results from spherical CNNs such as \cite{esteves:2020:spin}, as well as a 2D-planar CNN used for reference therein. When using the fully equivariant coefficients (REG), accuracy drops from $96.23$ to $92.3(1)$ from NR/NR to R/R,\footnote{NR/NR = train and test at fixed (non-rotated) orientation.  R/R = train and test at multiple orientations.  NR/R = train at fixed orientation, test at random orientations.} indicating the small drop in performance from the model having to account for all possible image angles. In contrast, an accuracy of only $50.2(5)$ was obtained for NR/R, showing the model trained at only one angle does not generalize well. The behavior of ISO is suppressed in magnitude but qualitatively similar, also showing a drop from NR/NR ($93.5$) to R/R ($92.10(5)$). Ideally, NR/NR and R/R would be identical for ISO, since no orientation information should be available to the LDA, so this decrease is an indication of slight imperfections in the rotational equivariance of the EqWS coefficients. This imperfection is reinforced by the accuracy of only $87.7(2)$ for NR/R. Given the residual imprint of the rectangular image grid, we also present results where images rotated at all three multiples of $180\degree/3$ were used for training. This represents a modest data augmentation motivated to suppress symmetry breaking from pixelation. In this augmented case, REG shows an NR/NR of $94.15$ (slightly higher than ISO trained at one angle) and NR/R of $88.4$, indicating a modest increase in robustness to rotations. However, one of the main results of this work is the equality of NR/NR, R/R, and NR/R for ISO in this case. All three cases have an accuracy of approximately $92.1$ and differ by less than $0.1\%$ (which is within 2$\sigma$ for the error shown).\footnote{Repeating the EqWS+LDA entries in the table using Lanczos interpolation for both image resizing and rotation leads to order $\sim 0.1\%$ increased mean classification accuracy and no significant change in the stability result between columns.}

This level of equality is not present in even state-of-the-art spherical CNNs, a subset of which are listed in the table for comparison. The best equality and overall accuracy was achieved by \cite{esteves:2020:spin}, showing a $0.4\%$ decrease in accuracy between NR/NR and NR/R, which was a major improvement over the original spherical CNN paper \cite{cohen:2018:spherical} showing a $2.19\%$ decrease. The errors shown for entries in this work are calculated as the standard deviation of classifications for $7$ distinct sets of (10,000) random rotations of the test images; the errors shown for \cite{esteves:2020:spin} are from different random initializations of network weights. Of course, we do not achieve overall accuracy competitive with these CNNs.\footnote{Of course, an algorithm that does not depend on the images, having an accuracy of $10\%$, can be perfectly invariant to rotations of the test images. We believe that our performance, while not state-of-the-art, is sufficient to support rotationally invariant learning.} However, the classification here required no training, and we have not optimized performance by systematic architecture optimization search ($c$, $w$, $L$, $m$).

\begin{figure}[ht!]
\centering
\includegraphics[width=\linewidth]{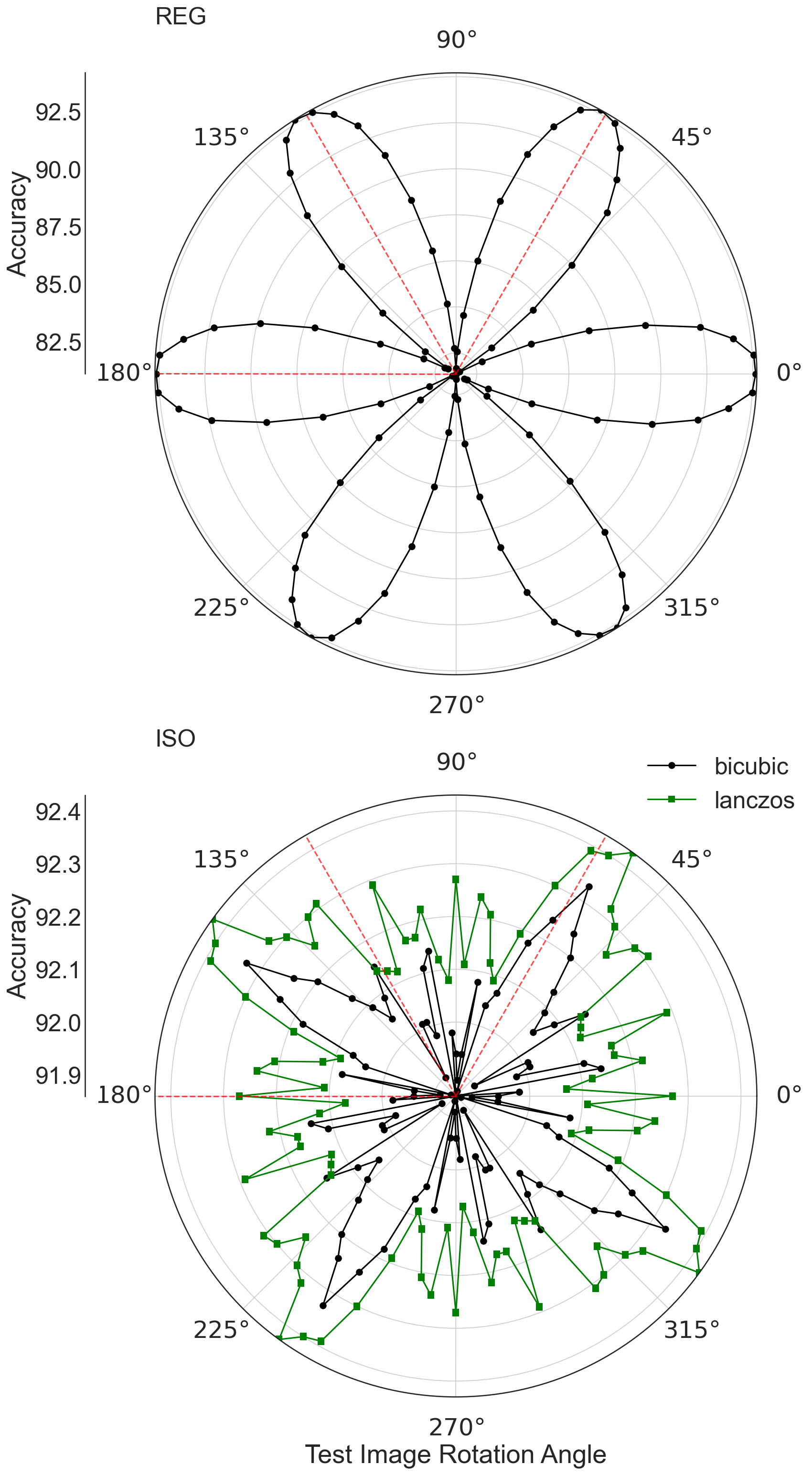}
\caption{Accuracy of EqWS+LDA on MNIST trained at three angles (all multiples of $180\degree/3$) on test images rotated at a uniform sampling of angles comparing using all coefficients (REG, \emph{top}) and the rotationally invariant reduction (ISO, \emph{bottom}). Note the difference in radial scale between the plots, which shows a two order-of-magnitude increase in rotational invariance (decrease in fluctuation amplitude). For the ISO case, the result is replicated using a Lanczos interpolation for both image resizing and rotation, illustrating that part of the residual, symmetry-breaking results from imperfect interpolation on the grid.
}
\label{fig:RotInvar}
\end{figure}

While a comparison of NR/NR and NR/R indicates how well a model generalizes, it does not indicate what angle dependence, if any, remains. To do this, we use a model trained at all multiples of $180\degree/3$ (three angles) while uniformly sampling the rotation angle of the test images. For REG, the accuracy smoothly oscillates up to $12\%$ as the test angle moves farther away from a train angle (Figure \ref{fig:RotInvar}, top). For ISO, the accuracy varies only at the $0.1\%$ level, a two order-of-magnitude improvement (Figure \ref{fig:RotInvar}, bottom). By looking at the accuracy as a function of angle, we can see that the remaining oscillations in the ISO coefficient accuracy have an approximate $90\degree$ symmetry. This is likely a result of the residual pixelation effects (see $j = 1 , 6$ in Figure \ref{fig:FinkletFilterBank} and Appendix \ref{sec:CoeffStab}) and imperfections in the bi-cubic interpolation implementing rotations. To support this, we repeated the experiment using a Lanczos interpolation, which better approximates an exact sinc interpolation, for both image resizing and rotation (Figure \ref{fig:RotInvar}, bottom). Improving the interpolation leads to a qualitatively more symmetric accuracy curve as a function of test-image rotation angle, a small increase in the mean accuracy ($\sim 0.2\%$), and small decrease in the standard deviation of the accuracy ($\sim 0.02\%$).\footnote{While we found Lanczos interpolation keeps the total image power an order of magnitude more stable than bi-cubic interpolation when implementing rotations, we benchmarked primarily with bi-cubic interpolations because this is more common and will more likely be implemented by the community. In contrast, using bi-linear interpolation with rotations suppresses power at large wavenumber, violating rotation invariance.} For applications requiring stronger invariance, stricter cuts on $j$ and wider wavelets in both the radial and angular directions can be used to mitigate these effects (see Appendix \ref{sec:CoeffStab}). We find this test even more satisfying than the usual test of training on one orientation and testing on random rotations because the symmetry of the coefficients is manifest in the accuracy response.

\begin{figure}[t!]
\centering
\includegraphics[width=\linewidth]{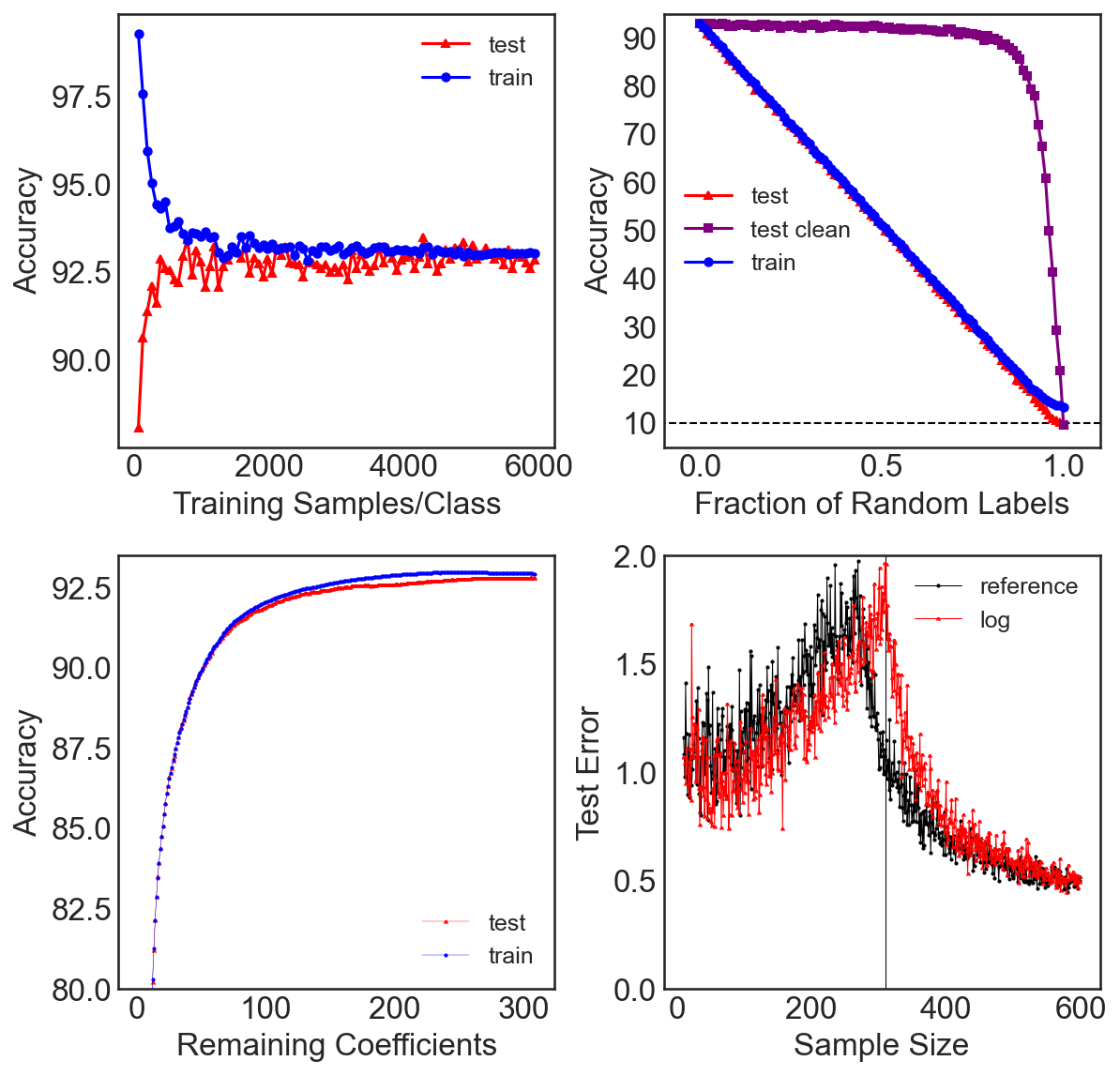}
\caption{\emph{Top Left:} Accuracy of MNIST classification by EqWS-ISO+LDA as a function of the number of training samples/class. Test and train accuracy are shown as red triangles and blue circles, respectively. \emph{Top Right:} Accuracy of MNIST classification by EqWS-ISO+LDA as a function of the fraction of false labels. Test (red triangle) has the same fraction of false labels as train, while test clean (purple squares) have all true labels. Both upper panels show that EqWS-ISO+LDA is robust to overfitting. \emph{Bottom Left:} Accuracy of MNIST classification by EqWS-ISO+LDA as a function of the number of coefficients remaining when coefficients are thrown out sequentially to maximize train accuracy. Test and train accuracy are shown as red triangles and blue circles, respectively; train accuracy increases above that at $310$ for the first $100$ coefficients throw-outs because we are explicitly maximizing the train accuracy. \emph{Bottom Right:} Accuracy of MNIST parity classification by EqWS-ISO+linear regression as a function of the total number of training samples. Accuracy from regression using the untransformed and log-transformed coefficients are shown as black circles and red triangles, respectively.
}
\label{fig:MLQA}
\end{figure}

In order to characterise how well EqWS+LDA generalizes and how robust it is, we first decrease the number of training samples per class to observe the onset of over-fitting and a large generalization gap (Figure \ref{fig:MLQA}, top left). At each point, we take a random test-train split with a unique seed. The test is always $10 \%$ of the combined MNIST set ($70,000$ images) and train is a random subset of the desired size. The accuracy remains fairly constant down to $\sim 1000$ training samples/class before diverging. Note that the steep change in accuracy occurs where the number of training samples/class is close to the dimension of the EqWS coefficients ($310$, because here we use ISO). Note that the variations in the test accuracy due to different random training subsets, even in the limit of a large number of training samples/class, are larger than the residual, angle-dependent fluctuations in accuracy in Figure \ref{fig:RotInvar}, bottom. This further validates that we have achieved a practical level of rotational invariance. Similarly, we perform MNIST classification where a variable fraction of the labels (both test and train) are randomized (Figure \ref{fig:MLQA}, top right). Here we again take a random test-train split with a unique seed, splitting $0.8/0.2$ train/test. A linear decrease in accuracy with no generalization gap is observed until above $90\%$ of the labels are randomized. When computing test accuracy on true labels for a model trained with fractionally randomized labels (purple squares), the test accuracy remains above $90\%$ until $80\%$ of train labels are randomized. This suggests that EqWS+LDA (ISO) is fairly robust to overfitting. 

To probe the dimensionality of the EqWS feature space, we sequentially eliminated the coefficient dimension that maximized the train accuracy, randomly selecting to break a tie. This process was repeated until only nine dimensions were left, those required to contain the class means for LDA (Figure \ref{fig:MLQA}, bottom left). Because we are optimizing for train accuracy, the train accuracy increases and exceeds the initial value at $310$ during the first $\sim 100$ steps. This increase is indicative of specialization to the subset of coefficients which perform best on this train set and is not transferable to other train sets. We present the average of $10$ random test-train splits in the figure to show that this specialization is in general possible. The test and train accuracy then agree again as both decrease rapidly below $50$ remaining coefficients. Note that over $65\%$ of the coefficients can be removed without reducing the test or train accuracy even $1\%$. This suggests there may be better dimension reductions beyond the ISO reduction explored here.

We also investigate the common practice of applying a log transform to the scattering coefficients, which has been shown to be useful for classification. To simplify the problem, we preform linear regression on the parity MNIST problem, classifying a coefficient as even or odd, so as to consider only a binary classification. The test error for regression on the untransformed and log-transformed coefficients as a function of the number of training samples (Figure \ref{fig:MLQA}, bottom right). The vertical line in the plot indicates when the number of samples is equal to the number of coefficients ($310$). The shift of the peak in the test error\footnote{The usual double descent feature.} indicates that the log transform regularizes the matrix inversion used in the linear regression, making the matrix closer to full rank. While we cannot conclude that the log transformation makes linear classification in general easier for all problems, this regularization and descriptive use of more dimensions of the feature vector is promising.

\subsection{Rotational Equivariance} \label{sec:RotEq}

We can leverage the smooth equivariance of the full set of coefficients to estimate the rotation angle of a given digit after training on a few angles, though we can hope only to estimate rotation angles $[0,180\degree)$ because of the modulus in EqWS.\footnote{One could attempt to further leverage the equivariance of the coefficients by creating synthetic training coefficients where the equivariant permutation Eq. \ref{eq:equivPermute} is applied to the coefficients and the angle labels for the MNIST digits are modified to $s \times 180\degree/L$, but we find that this does not improve angle estimation in practice.} Here we only pad the MNIST digits in a $64 \times 64$ image and interpolate to a $128 \times 128$ image so we can sample a large range of angles. We compute the EqWS coefficients for all images in the test or train set at $24$ train angles or $48$ test angles $[0,180\degree)$. We then learn the digit class at fixed angle, the angle of rotation for a fixed digit class, or both the digit class and angle simultaneously.

\begin{figure}[b!]
\centering
\includegraphics[width=\linewidth]{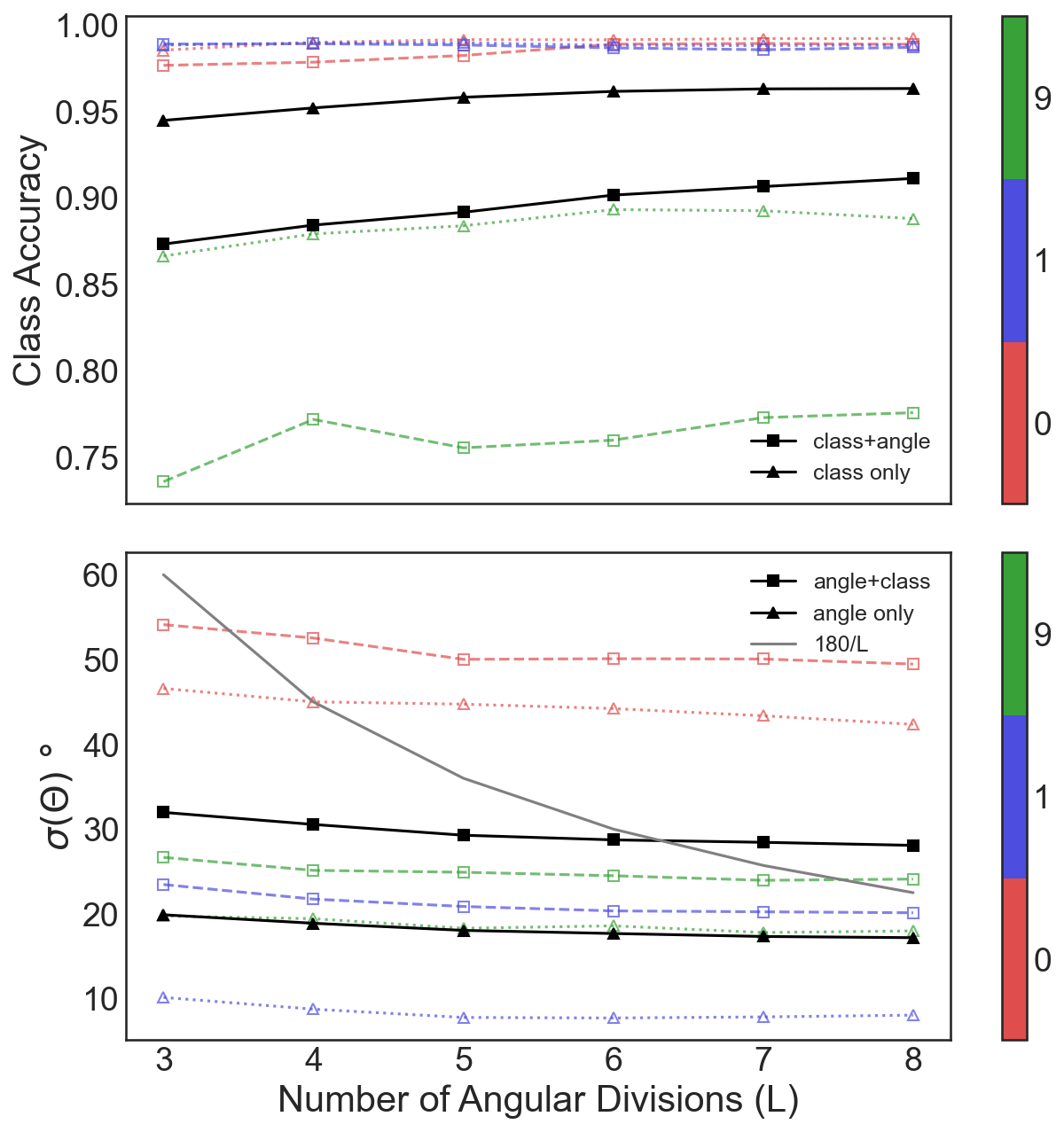}
\caption{\emph{Top:} Mean accuracy of digit classification when simultaneously learning the digit class and angle of rotation (solid line squares) and when only learning the digit class at fixed angles (solid line triangle, average of $24$ train angles). Accuracy for the extremal classes are displayed separately for ``class+angle'' (dashed square) and ``class only'' (dotted triangle). \emph{Bottom:} Standard deviation of the true minus predicted angle of rotation when simultaneously learning the digit class and angle of rotation (solid line squares) and when only learning the angle of known classes (solid line triangle). Accuracy for the extremal classes are displayed separately for ``angle+class'' (dashed square) and ``angle only'' (dotted triangle). Color indicates class as $0$, $1$, or $9$, as indicated by the color bar. 
}
\label{fig:LSeries}
\end{figure}

We learn both the digit class and rotation angle simultaneously, training using a varying subset size of the $24$ train angles computed above. The accuracy plateaus by $12$ train angles (separation of 15 $\degree$), so we present results with this number.\footnote{The only metric that does not plateau as a function of the number of train angles is the standard deviation of the digit class accuracy as a function of angle, which follows expectations for data augmentation and follows from Figure \ref{fig:RotInvar}.} With regards to learning the class, we achieve $>90\%$ average accuracy over all classes for $L=8$. As the number of angular bins used to compute the EqWS coefficients increases, the average accuracy also increases (solid line squares, Figure \ref{fig:LSeries}, top). We compare this to the case where we learn the digit class at fixed angle, which is similar to the NR/NR case in Table \ref{table:NR_R}\footnote{The ``class+angle'' case is akin to the NR/R with data augmentation case for the digit classes in Table \ref{table:NR_R} and the "class only" case is like NR/NR at different angles where the average over angles is reported in Figure \ref{fig:LSeries}. We find few percent, class-dependent fluctuations as a function of angle within the "class only" data which are largest for $3$, $6$, and $9$. We attribute these fluctuations to an interplay of imperfect interpolation and wavelet pixelation.}, except that we are using one-hot linear regression here (solid line triangles, Figure \ref{fig:LSeries}). As expected, when the regression only has to predict the class, higher accuracy is achieved ($\sim 96\%$). The increase in accuracy as a function of $L$ is far slower, and appears to saturate near $L=8$. This suggests that the $L$ dependence of the simultaneous task does not derive from increased angular resolution, improving class labeling for digits at fixed angles. We also show the accuracy of the extremal classes, $0$, $1$, and $9$ in Figure \ref{fig:LSeries}. There is a very small gap between ``class+angle'' and ``class'' for $0$ and $1$ which are identified with high accuracy in both cases. In contrast, there is a $\sim 10\%$ gap for $9$ where the orientation is informative for classification.

To learn the angle of rotation of an image, we use linear regression to independently predict $\cos{\theta}$ and $\sin{\theta}$ and estimate $\theta$ as the $\arctan$ of the cosine and sine predicted values. Here $\theta$ is actually twice the angle of rotation since EqWS has $180\degree$ symmetry. We quantify the ability to estimate this angle as the standard deviation ($\sigma_\theta$) of $\theta-\theta_{\rm{predicted}}$ and present the results in Figure \ref{fig:LSeries} (bottom). The mean $\sigma_\theta$ for all classes is $\sim 30 \degree$, decreasing as $L$ increases. For comparison, we show the mean $\sigma_\theta$ when the class is fixed and only the angle is learned (solid line triangles, Figure \ref{fig:LSeries}) where the mean $\sigma_\theta$ improves by $\sim 10 \degree$. Just as with the digit classes, we show the extremal classes, $0$, $1$, and $9$. Unsurprisingly, it is most difficult to predict the angle of rotation of $0$, which is most nearly rotationally symmetric, and easiest to estimate the angle of rotation for $1$, which, similar to the rod tests, primarily activates a single wavevector. The largest gap between ``class+angle'' and ``angle'' was for $1$, which has a $\sigma_\theta$ of only $10 \degree$. As another point of comparison, we show $180/L$, which might be the naive limit of angular resolution for $L$ angular divisions. For $L<8$, the "angle only" line is below $180/L$ and $\sigma_\theta$ decreases much more slowly with $L$ than $1/L$. We take this as support for the angle being easily accessible not only by which $\ell$ coefficient is maximal, but from the relative magnitude of each $\ell$ during the continuous trade-off in power between reference angles as a function of rotation (Figure \ref{fig:RotEquiv}). We suspect that the value of the plateau of $\sigma_\theta$ in part derives from the intrinsic scatter in the as-written angle of the digits, in addition to limitations arising from EqWS.\footnote{It is difficult to compare to literature since there is no single agreed-upon metric for angle estimation. This is complicated by the fact that the distribution of incorrect angle assignments has peaks at $90\degree$ and $180\degree$ and is thus not well described by its mean or standard deviation. However, our results are broadly consistent with other CNN-based approaches \cite{blankrot:rotmnist}.} The combination of the plateau in $\sigma_\theta$ for both "angle+class only" and "angle only" and the plateau in class accuracy for "class only" suggests that the $L$ dependence of the simultaneous digit classification is unique to learning the distribution of digit classes at different angles and does not simply derive from improved angle estimation or class prediction (at a fixed angle).

\begin{figure}[t!]
\centering
\includegraphics[width=\linewidth]{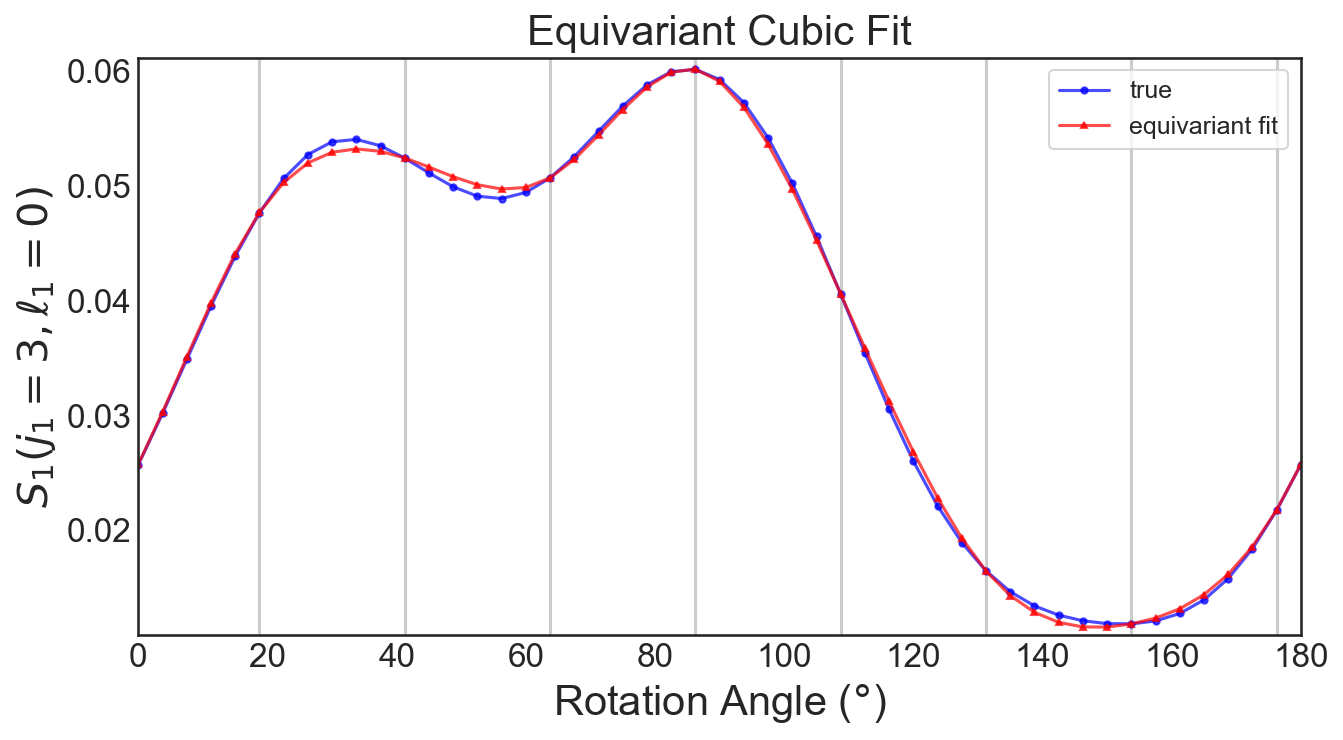}
\caption{Angle dependence of the $S_1(j_1=3,\ell_1=0)$ for an example digit (``2'') from the MNIST test set. True coefficients computed on image rotated by (pixel-wise) bi-cubic interpolation shown as blue line with circles. Grey vertical lines denote the coefficient values known by equivariant permutation from the EqWS coefficients at a single image rotation angle. Cubic spline interpolation of the $L=8$ equivariant fixed points shown in red (line triangles) closely matches the true coefficient angle dependence, allowing approximate continuous equivariance as a result of the smooth angular dependence.
}
\label{fig:EquivarFit}
\end{figure}

By combining the relatively smooth, continuous, angular dependence and the discrete $L$-fold equivariance of the EqWS coefficients, we can predict the EqWS coefficients at arbitrary angle with minor errors, achieving approximate equivariance with respect to continuous rotations. To demonstrate this, Figure \ref{fig:EquivarFit} shows the $S_1(j_1=3,\ell_1=0)$ coefficient as a function of rotation angle for a test image from MNIST, a ``2'' (blue line circles), using the resizing and interpolations described above. Given the image at a single orientation, we can predict the value of $S_1(j_1=3,\ell_1=0)$ at $L=8$ discrete rotations by permutation (Equation \ref{eq:equivPermute}, gray lines Figure \ref{fig:EquivarFit}). Given the smooth angular dependence, we can perform a simple bi-cubic spline fit (with periodic boundary conditions) and predict the $S_1(j_1=3,\ell_1=0)$ coefficient for arbitrary rotation angles. The average error over the $48$ test angles was $1 \times 10^{-3.4}$ for the $S_1$ and $1 \times 10^{-4.4}$ for the $S_2$ coefficients, which are conveniently in units of the fractional image power. This error decreases as a function of $L$ as expected, but does so slowly (less than an order of magnitude from $L=3$ to $L=8$).


\section{Applications/Benchmarks} \label{sec:Benchmarks}

In this work, we focus on maximizing equivariance and so we do not attempt a full architecture optimization ($c$, $w$, $L$, $m$) to maximize accuracy. However, we wish to establish a baseline on more difficult datasets such as EMNIST, CIFAR-10, and CIFAR-100. EMNIST is extended version of MNIST, which includes handwritten letters in addition to handwritten digits \cite{cohen:2017:emnist}. The EMNIST digits are padded in a $64 \times 64$ image and interpolated to a $128 \times 128$ image. CIFAR-10 is a set of 10 mutually exclusive classes of $32 \times 32$ pixel color images of objects and animals \cite{krizhevsky:2009:learning}. CIFAR-100 has 100 classes of the same image format as CIFAR-10, but those 100 classes are grouped into twenty super-class labels that contain five of the 100 fine-label classes. Before computing the wavelet scattering coefficients, we up-sample the CIFAR images to $64 \times 64$, apodize to the image mean, and then pad the images to $128 \times 128$. In order to use the definition of the wavelet scattering coefficients above, we convert the images to monochrome using $0.299 \times R + 0.587 \times G + 0.114 \times B$. 

To define the wavelet scattering network on color images, we could take the simplest possible extension, adding an additional channel index and obtaining three times as many coefficients. We do this for the first-order coefficients, but replace the second-order coefficients by defining a cross term to pick up correlations between color channels, where the multiplication below is taken element-wise in real space. A more thorough comparison of this form of second-order coefficient and the usual one, both defined on a monochrome image, is warranted and will appear in later work. We view this form of second-order coupling as the simplest way to obtain correlations between the color channels and use it here for benchmarking.\footnote{To our knowledge, this sort of color-channel coupling is novel for wavelet scattering networks. Most wavelet scattering work on color images treats each channel separately or combines them into a single channel \cite{oyallon2015deep,cotter2017visualizing}.}

\begin{ceqn}
\begin{align}
S_2^{C_1 \times C_2}(j_1,\ell_1,j_2,\ell_2) & \\
 = \int |\bar I_{C_1} & \star \psi_{j_1,\ell_1}| \times |\bar I_{C_2} \star \psi_{j_2,\ell_2}|(\vec x) \;d^2\vec x \nonumber
\end{align}
\end{ceqn}
For coefficients with $C_1=C_2$, $j_1=j_2$, and $\ell_1=\ell_2$, the above $S_2$ coefficients conveniently reduce to the usual $S_1$ coefficients for $p=2$, so there is no need to compute those separately.

For EMNIST, the performance of EqWS+LDA surpasses a simple linear classifier and is comparable to the three-layer extreme learning machine (ELM) presented in the EMNIST release paper, though EqWS+LDA requires no training \cite{cohen:2017:emnist}. Monochrome EqWS+LDA performs similarly to a single hidden layer neural network applied to the pixel space representation with logistic regression on the outputs ($51.53\%$) presented in the CIFAR-10 release \cite{krizhevsky:2009:learning}. However, the accuracy of monochrome, rotationally invariant EqWS+LDA is more comparable to simple logistic regression on the pixel space representation ($41.13\%$). By extending to cross-color-channel scattering coefficients, the performance of the ISO coefficients is most improved, increasing from $42\%$ to $49\%$ for CIFAR-10 and from $16\%$ to $34\%$ for CIFAR-100. For the REG coefficients, a less than $1\%$ increase on CIFAR-10 and $15\%$ increase on CIFAR-100 was obtained. Logarithmic transform and normalization of the EqWS coefficients, as well as using a different color space for the images, can improve accuracy on CIFAR $\sim 3-5\%$ (see Appendix \ref{sec:ExtApp}.)

\begin{table}[ht]
\centering
\caption{Accuracy of EqWS+LDA on Benchmark Datasets}
\begin{tabular}{l|cc}
\toprule
Dataset & REG & ISO \\

\midrule
EMNIST Digits & $96.39$ & $92.56$ \\
EMNIST Letters & $83.59$ & $69.61$ \\
EMNIST Balanced & $77.23$ & $65.72$ \\
CIFAR-10 Grey & $53.53$ & $42.39$ \\
CIFAR-10 RGB & $54.00$ & $48.58$ \\
CIFAR-100 Grey & $20.36$ & $16.27$ \\
CIFAR-100 RGB & $35.29$ & $34.03$ \\
\bottomrule
\end{tabular}
\end{table}

In this work, we have shown that there may be a yet lower dimensional reduction of the wavelet scattering coefficients, beyond a simple ISO reduction (sum over $\ell$ indices), which could retain almost all of the information useful for classification (on the specific MNIST task considered). Another reduction of the standard wavelet scattering transform (WST) has been proposed, known as RWST (see Appendix \ref{sec:ExtApp}, and figure therein). This reduction fits the angular dependence of the first- and second-order coefficients by one cosine and three cosine terms, respectively, each with an overall constant for centering. This reduction retains some anisotropic angular information in contrast to ISO.

In order to compare different representations of the wavelet scattering coefficients, we compare ISO and RWST reductions on an isotropic classification problem. To do this, we use the original definition of the wavelet scattering coefficients ($p=1$) on which RWST is defined, and compare this to the ISO sum over $\ell$ indices of those same coefficients.\footnote{We use the publicly available \textsc{Kymatio} package with which RWST was released for this comparison.} We consider a dataset of eight classes of magnetohydrodynamic (MHD) simulations characterized by two different dimensionless numbers on which RWST was previously benchmarked and which is of interest to the astrophysics community \cite{saydjari2020classification}. The ISO reduction ($75\%$) outperforms RWST ($69\%$) at the classification task while using Morlet wavelets which are not exactly rotationally invariant. The constant offsets which are fit in RWST act approximately like a mean over the $\ell$ indices, which is just a normalized sum. Limiting the RWST coefficients to only these constant coefficients (but fitting the entire functional form) results in an accuracy of ($78\%$). In comparison, EqWS-ISO+LDA with the default filter bank results in an accuracy of $82\%$.

For this isotropic task at least, it appears that the isotropic reductions are simpler and sufficient. In general, we might desire more descriptive coefficient representations such as REG and or RWST to perform as well at isotropic tasks as ISO, though undertraining and overfitting often stand in the way. Other coefficient reductions which retain anisotropic information and their application to classification and regression tasks which benefit from anisotropic information, such as the direction of a magnetic field in MHD simulations, are under study. However, the MHD simulations studied here are different enough to be distinguished by only first-order EqWS coefficients with an accuracy of $81\%$. Ongoing work is also exploring a more thorough analysis of which MHD classification and regression problems require or benefit from second-order coefficients. 


\section{Conclusion}

We introduced a new set of wavelets (triglets) with a modified scattering network, EqWS, in order to optimize the rotational equivariance/invariance and translation invariance of wavelet scattering statistics. In the Fourier domain, the wavelets point-wise sum to one within the Nyquist disc, are dynamically adjusted to be well-sampled at small $k_r$ (large spatial scales), and are nowhere constant where nonzero. Code to implement EqWS is released with this work and takes advantage of wavelet sparsity and pooling under the $p=2$ power. We show that EqWS is translation invariant and rotation equivariant. Each angular bin is equivalent up to sampling effects at the largest and smallest scales. While we show that the EqWS coefficients at a given scale peak as a function of the object scale, we caution that scale equivariance requires sufficient margin on both small scales (the PSF must be well sampled at all rescalings of interest) and large scales (the information must not spill over the image boundaries). This may require both upsampling and padding the input image. 

We studied EqWS and its isotropic reduction with simple linear methods (LDA and linear regression) on MNIST. We find almost no difference between training on randomly rotated images and on images rotated at one of three angles when testing on images with random rotations. Residual fluctuations in accuracy as a function of angle suggest a symmetry-breaking term with fourfold symmetry, likely resulting from the discretized image grid. We leveraged the smooth equivariance of the EqWS coefficients to simultaneously learn the class and rotation angle of MNIST digits, observing very little dependence on the number of angular divisions ($L$). We further predict the continuous angular dependence of a scattering coefficient from a single image and angle using the coefficients at the same scale, but different angular bins. We also benchmark on EMNIST and CIFAR-10/100, introducing a new second-order, cross-color-channel coupling term. Revisiting prior wavelet scattering coefficient dimension reductions in the context of an isotropic classification of magneto-hydrodynamic simulations supports a simple isotropic sum such as the one used here.


\section{Code and Data Availability} \label{sec:dataavil}

We release code to implement EqWS in \textsc{Julia} in a public GitHub  \href{https://github.com/andrew-saydjari/EqWS.jl}{\textbf{EqWS.jl}}. Data products associated with the paper are publicly available at \url{https://doi.org/10.5281/zenodo.4686088} (48 GB). This includes computed EqWS coefficients, code for preprocessing each dataset, and code to reproduce all EqWS coefficients. A more ML friendly version of the MHD dataset introduced in \cite{saydjari2020classification} is available \href{https://faun.rc.fas.harvard.edu/saydjari/EqWS/MHD_ML/}{\textbf{here}}. Also included are \textsc{Jupyter} notebooks containing code to reproduce all figures in the text, some minimal working examples, and how to run these computations on a cluster. 

We used many publicly available codes, including \textsc{Python} packages: 
\textsc{HDF5} \cite{collette_python_hdf5_2014}, 
\textsc{ipython} \cite{Perez:2007:CSE:}, 
\textsc{kymatio} \cite{Andreux:2018:arXiv:}, 
\textsc{matplotlib} \cite{Hunter:2007:CSE:}, 
\textsc{numpy} \cite{vanderWalt:2011:CSE:},
\textsc{scipy} \cite{Virtanen:2020:NatMe:}, 
\textsc{scikit-learn} \cite{Pedregosa:2012:arXiv:}, 
\textsc{scikit-image} \cite{vanderWalt:2014:arXiv:},
\textsc{dynesty} \cite{Speagle:2020:MNRAS:} 
and \textsc{Julia} \cite{bezanson2017julia} packages:
FFTW.jl \cite{FFTW.jl-2005},
AbstractFFTs.jl,
Colors.jl,
DSP.jl,
FITSIO.jl,
HDF5.jl,
IJULIA.jl,
MLDatasets.jl,
and would like to acknowledge their developers.


\section*{Acknowledgment}

A.S. gratefully acknowledges support by a National Science Foundation Graduate Research Fellowship (DGE-1745303). D.F. acknowledges support by NSF grant AST-1614941, “Exploring the Galaxy: 3-Dimensional Structure and Stellar Streams.” We acknowledge Cammarata et al. \cite{cammarata:2020:curve}, whose figure pushed us to improve Figure \ref{fig:RotEquiv}, and found this up-to-date literature \href{https://github.com/Chen-Cai-OSU/awesome-equivariant-network#equivariance-and-Group-convolution}{\textbf{list}} highly helpful. We thank Taco Cohen, Josh Speagle, Catherine Zucker, Core Francisco Park, Justina R. Yang and Nayantara Mudur for helpful discussions. We thank Erwan Allys for helpful discussions and feedback on a draft of this work. A.S. acknowledges Sophia S\'{a}nchez-Maes for helpful discussions and much support. Computations in this paper were run on the FASRC Cannon cluster supported by the FAS Division of Science Research Computing Group at Harvard University.


\appendices
\section{Wavelet Comparison}\label{sec:MorletOptim}

\begin{figure}[t!]
\centering
\includegraphics[width=\linewidth]{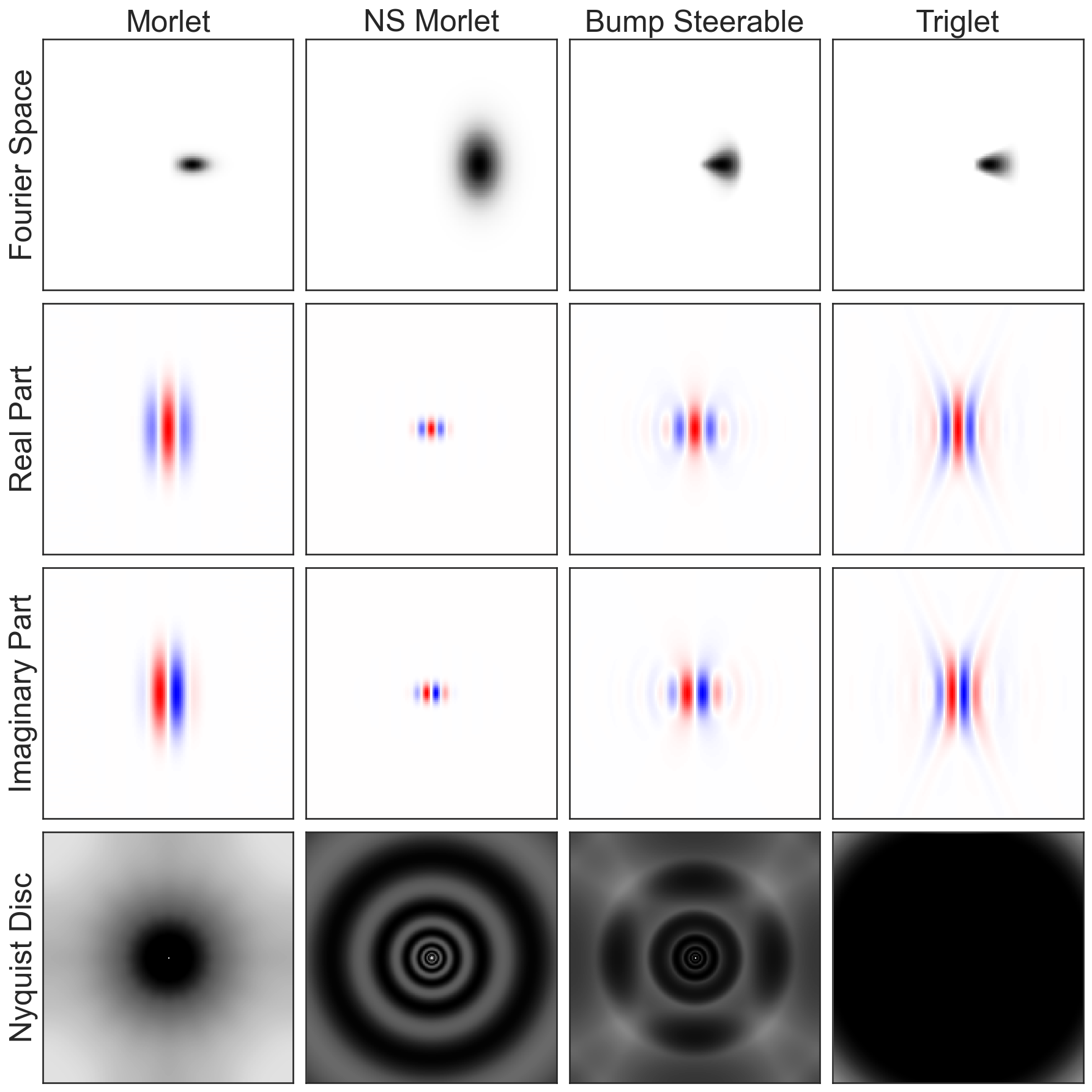}
\caption{Comparison of selected wavelets at $j=3$ and $l=0$ for ``Morlet'', ``NS Morlet'', ``Bump Steerable'', and ``Triglets.'' Fourier space plots are min-max normalized with dark colors representing high values. Real-space plots are symmetrically min-max normalized around zero, with red positive and blue negative. The real and Fourier space plots are cropped to $128 \times 128$. The Nyquist disc coverage for ``NS Morlet'' and ``triglets'' is the sum of the squares, but for ``Morlet'' and ``Bump Steerable'' it is the sum of the absolute value of the wavelets in Fourier space.
}
\label{fig:WaveletCompare}
\end{figure}

We show selected wavelets at $j=3$ and $\ell=0$ side-by-side in Figure \ref{fig:WaveletCompare} for visual comparison. The Morlet wavelets use geometrical parameters found in the \textsc{Kymatio} filter bank. The ``NS Morlet'' are Morlet wavelets optimized using nested sampling, implemented in \textsc{dynesty}, so that the sum of squares of the wavelets was uniform across the Nyquist disc. The ``Bump Steerable'' wavelets are those implemented in \textsc{ScatNet} and are of the class being used widely in an extension of the wavelet scattering transform (WST) called wavelet phase harmonics (WPH) \cite{mallat2020phase}. The ``triglets'' are the wavelets we introduced in the main text. The real and Fourier-space plots are cropped to $128 \times 128$ to aid in visualization. To illustrate how uniformly the wavelets cover Fourier space, we plot the sum of the wavelets squared for the entire filter bank for the ``NS Morlet'' and ``triglets.'' The best comparison for the ``Morlet'' and ``Bump Steerable'' wavelets is to plot the sum of the absolute value of those wavelets since they are used with the original definition of the wavelet scattering transform which pools under the modulus (i.e. $p=1$).

The ``NS Morlet'' wavelets are parameterized by $a, b, c$ and $d$ on which we have place uniform priors from $0$ to [$2, 2, 4, 2$].

\begin{ceqn}
\begin{align}\label{eq:WST_2D_wavelet}
\widehat{\psi_{j\ell}}(x) & = d \times (e^{a 2^{j} ix}-\beta)e^{-x^2/(2\sigma^2)}
\end{align}
\end{ceqn}

Here $\beta$ is a normalization factor \cite{Ashmead:2010:arXiv:}, which depends on $a, b, c$ chosen such that $\psi$ has a null average. Since we are working in 2D, the inverse variance $1/\sigma^2$ of the Gaussian window is an inverse covariance matrix. We choose a diagonal covariance matrix with $\sigma_{yy} = c \times \sigma_{xx}$ (where $x$ is the direction of the plane wave) and let $\sigma_{yy}$ = $b \times 2^j$.

The nested sampling was performed with a Gaussian log-likelihood (variance $10^{-7}$) on the mean-squared error (MSE) between the sum of the squares of the wavelets and the unit disc with radius $128$. Note that in the optimization and Figure \ref{fig:WaveletCompare} $j \in [0,8]$ even though we exclude $j=0$ and $j=8$ afterward since those wavelets are poorly sampled either in real or Fourier space. The optimized parameters for this parameterization and a few others we explored are available in Sec. \ref{sec:dataavil} along with detailed implementation choices for the nested sampling.

\section{Computational Cost/Accuracy}\label{sec:CompCost}
To illustrate the trade-off between equivariance and computational cost that arises from making sure images are well-sampled in real and Fourier space, we show the accuracy of EqWS-ISO+LDA on MNIST for various amounts of image padding and upsampling. For this test, we rotate the train images by $3$ equally spaced angles and test images by $50$ equally spaced angles (from $0$ to $180 \degree$) and report the mean and standard deviation of the test accuracy over the test angles using LDA basis trained on the $3$-angle augmented training set (see Table \ref{table:delta}). For (padding, upsampling) $= (0,0)$, the $28 \times 28$ MNIST images are embedded in the smallest dyadic image $32 \times 32$ without any further processing before the EqWS coefficients are computed. For (padding, upsampling) $= (q,r)$, the MNIST images are embedded in a $2^{5+q} \times 2^{5+q}$ image and interpolate to a $2^{5+q+r} \times 2^{5+q+r}$ image. The number of coefficients and computational cost (see Table \ref{table:compCostJL}) depend only on the image size $2^{5+q+r}$ since we fix the rest of the filter-bank parameters to the default values $L=8$, $w=2$, $c=1$, and $p=1$.

To quantify the stability of the isotropic coefficients formed from the EqWS coefficients, we report $\Delta$ (see Table \ref{table:delta}).

\begin{ceqn}
\begin{align}\label{eq:deltadef}
\Delta & = \frac{1}{N_{\rm{coeff}}N_{\rm{images}}}\sum_{i}{\sum_{c}{\sigma_{i,c}}}
\end{align}
\end{ceqn}

where $i$ indexes the set of test images, $10000$, $c$ indexes the different EqWS-ISO coefficients, $310$ for (padding, upsampling) $ = (2,1)$, and $\sigma_{i,c}$ is the standard deviation of the coefficient on that image over the $50$ equally spaced rotation angles tested. The unit of $\Delta$ is image power, which is useful for comparing the stability of the coefficients with the conservation of image power as a function of the interpolation scheme used for the rotation.

\begin{table}[ht!]
\centering
\caption{Mean and Standard Deviation of MNIST Accuracy achieved by EqWS+LDA with varying Padding and Upsampling}
\begin{tabular}{c|cccc}
\toprule

\multirow{2}{*}{Padding} & \multicolumn{4}{c}{Upsampling} \\
& $0$ & $1$ & $2$ & $3$ \\

\midrule
$0$ & $83.62 \pm 0.64$ & $84.78 \pm 0.62$ & $86.67 \pm 0.50$ & $87.40 \pm 0.40$ \\
$1$ & $90.46 \pm 0.21$ & $91.04 \pm 0.16$ & $92.00 \pm 0.14$ &  \\
$2$ & $91.61 \pm 0.12$ & $92.06 \pm 0.12$ & & \\
$3$ & $92.17 \pm 0.10$ & & & \\
\bottomrule
\end{tabular}
\label{table:muSigma}
\end{table}

\begin{table}[ht!]
\centering
\caption{Average EqWS Coefficient Variability ($\Delta$) on MNIST with varying Padding and Upsampling}
\begin{tabular}{c|cccc}
\toprule

\multirow{2}{*}{Padding} & \multicolumn{4}{c}{Upsampling} \\
& $0$ & $1$ & $2$ & $3$ \\

\midrule
$0$ & $2.7\times10^{-4}$ & $1.6\times10^{-4}$ & $1.1\times10^{-4}$ & $7.6\times10^{-5}$ \\
$1$ & $4.8\times10^{-5}$ & $2.9\times10^{-5}$ & $2.0\times10^{-5}$ &  \\
$2$ & $1.6\times10^{-5}$ & $8.0\times10^{-6}$ & & \\
$3$ & $8.1\times10^{-6}$ & & & \\
\bottomrule
\end{tabular}
\label{table:delta}
\end{table}

To compare our code to the most popular public scattering network code, \textsc{Kymatio}, we compute scattering coefficients on 2D images of size $2^{J^{\rm{im}}}$ pixels with $L=8$. Since the two codes choose the number of scales to compute differently, we compare the computation of the largest number of scales possible for a given image size in each code. We report the total computational time per call, the total number of coefficients computed for an image of that size, and the computational time per coefficient.

\begin{table}[hb!]
\centering
\caption{Computational cost for \textsc{Kymatio} and EqWS.jl}
\begin{tabular}{c|rr|rr|rr}
\toprule

\multirow{2}{*}{$J^{\rm{im}}$} & \multicolumn{2}{c|}{Time (core-ms)} & \multicolumn{2}{c|}{Number of Coeff} & \multicolumn{2}{c}{Time/Coeff (core-$\mu$s)}\\
& \multicolumn{1}{c}{EqWS} & \multicolumn{1}{c|}{WST} & \multicolumn{1}{c}{EqWS} & \multicolumn{1}{c|}{WST} & \multicolumn{1}{c}{EqWS} & \multicolumn{1}{c}{WST}\\

\midrule
$8$ & $270$ & $6300$ & $2452$ $(310)$ & $1857$ & $110$ & $3400$ \\
$7$ & $46$ & $1700$ & $1724$ $(219)$ & $1401$ & $27$ & $1200$ \\
$6$ & $13$ & $578$ & $1124$ $(144)$ & $1009$ & $12$ & $570$ \\
$5$ & $5.0$ & $233$ & $652$ $(85)$ & $681$ & $7.7$ & $340$ \\
$4$ & $2.5$ & $110$ & $308$ $(42)$ & $417$ & $8.1$ & $260$ \\
$3$ & $1.2$ & $51$ & $92$ $(15)$ & $217$ & $13.0$ & $240$ \\
\bottomrule
\end{tabular}
\label{table:compCostJL}
\end{table}

\textsc{Kymatio} excludes second-order coefficients with $j_1 > j_2$ on physical grounds \cite{Allys:2019:A&A:} and thus has to compute fewer coefficients. In EqWS.jl, we have not yet established that these coefficients are uninformative and instead retain all second order coefficients. The number of coefficients after the isotropic reduction of the EqWS coefficients are shown in parentheses. These experiments were executed on the FASRC Cannon cluster at Harvard University on a compute node with water-cooled Intel 24-core Platinum 8268 Cascade Lake CPUs with 192GB RAM running 64-bit CentOS 7. Cascade Lake cores have dual AVX-512 fused multiply-add (FMA) units. Code scaling is reported on a single core in a \textsc{Python} or \textsc{Julia} environment specified by the \textsc{yaml} file in Sec. \ref{sec:dataavil}. We also observe an additional speed-up of $\sim 10\times$ when both codes are parallelized naively in the respective languages, but do not include this in the benchmarking until more rigorous attempts are made to make the parallelized code comparable.

\section{Coefficient Stability}\label{sec:CoeffStab}

\subsection{Rod Test}\label{sec:CoeffStabRodTest}
\begin{figure}[hb!]
\centering
\includegraphics[width=\linewidth]{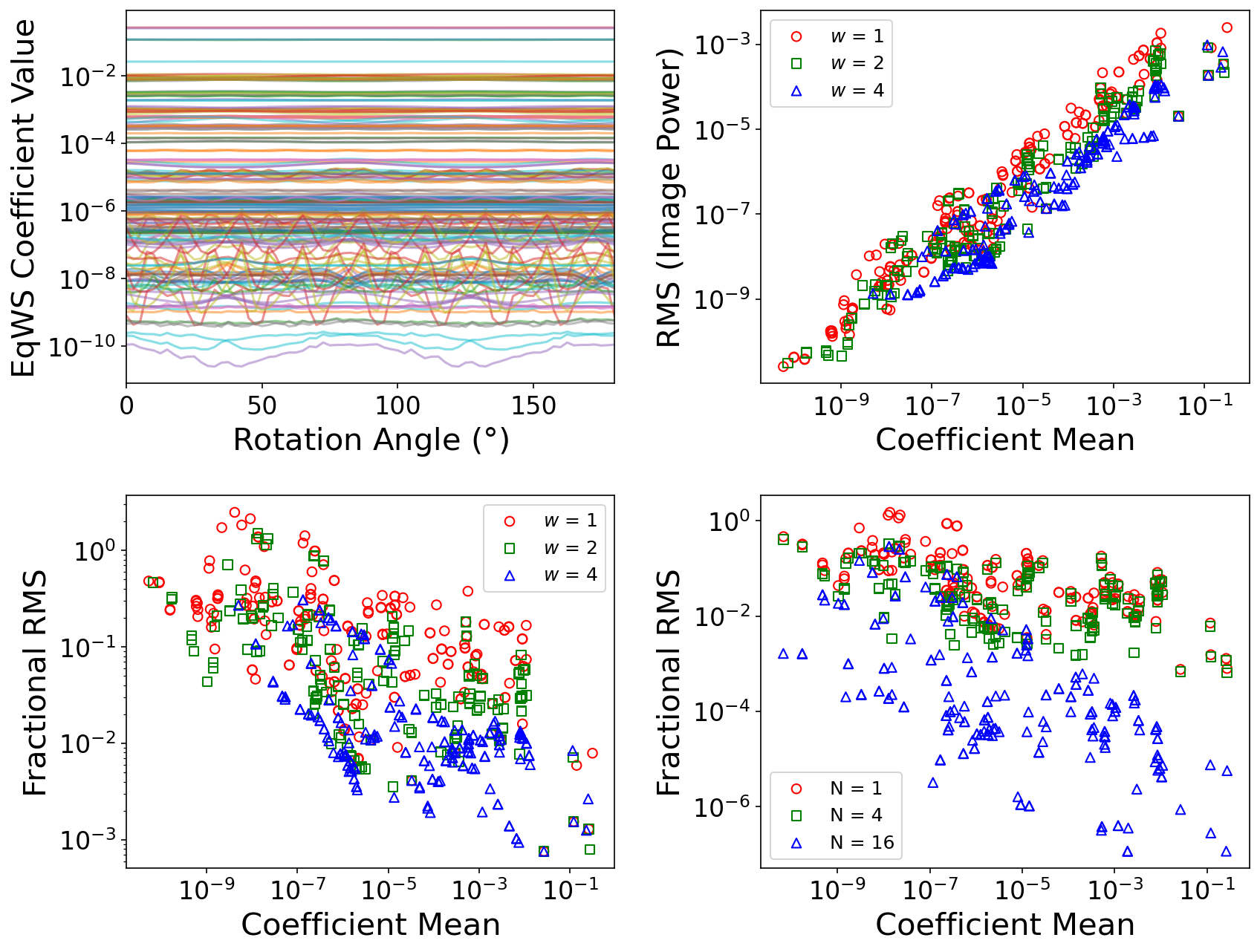}
\caption{\emph{Top left:} All $S_2$ coefficients for an image of a rod as a function of rotation angle. These coefficients are computed with standard filter bank parameters $w=2$, $c=1$, $p=1$, and $L=8$. Periodicity of oscillations informs symmetry-breaking terms. \emph{Top right:} RMS of each coefficient plotted versus the coefficient mean using wavelet filter banks with various angular filter widths ($w = 1, 2, 4$ as red circle, green square, blue triangle). \emph{Bottom left:} Same as \emph{top right} except showing the fractional RMS. \emph{Bottom right:} Fractional RMS of each coefficient plotted versus the coefficient mean where the estimate of each coefficient (which is isotropic) is a mean of coefficients obtained from N image rotations. Wider wavelets and averaging angular samples suppress the coefficient oscillations which indicate deviations from the coefficients being fully isotropic.
}
\label{fig:CoeffStable}
\end{figure}

We further investigate oscillations in the ISO EqWS coefficients, which represent deviations from being fully rotationally invariant. Similar to the main text, we use images of rods which are $128 \times 128$ pixels, FWHM $=8$, and $30$ pixels long. Fluctuations in coefficient values as a function of the rod angle is shown for $S2$ for all $j$ and $\ell$ indices in Figure \ref{fig:CoeffStable} (top left). The rotation angle is sampled every $2.5 \degree$. While this representation can be useful in identifying the periodicity of oscillations informing the origin of the symmetry breaking, it is difficult to interpret and compare. We instead show the RMS fluctuations of the coefficients (which are units of image power) as the images are rotated versus the mean value of that coefficient (Figure \ref{fig:CoeffStable}, top right). We then change the angular width of wavelets used from the default $w = 2$ to $w = 1$ and $4$. Widening the wavelets decreases the RMS as indicated by the relative vertical offsets between the centroid of the $3$ scatter-point distributions. Widening the wavelets also increases the minimum coefficient mean, which agrees with the intuition that wider wavelets have decreased angular sparsity with respect to the rod.

It is often useful to examine the fractional fluctuations of the coefficients, which has been previous practice in the equivariant CNN literature \cite{cohen:2018:spherical} (Figure \ref{fig:CoeffStable}, bottom left). For EqWS, the smallest coefficients have the largest fractional fluctuations, as can be seen qualitatively in Figure \ref{fig:CoeffStable} (top left) and by the negative slow in Figure \ref{fig:CoeffStable} (bottom left). Another approach to suppressing the oscillations of the ISO coefficients as a function of angle is to simply sample coefficient values at a few (N) angles and average. This is a form of training augmentation, but here the classifier is not trained on N times as many samples; it is only that N times as many coefficient computations enter into each coefficient reported to the classifier.\footnote{This averaging procedure changes the statistics of the coefficients of the train set relative to the test set, unless the same number N rotations of the test image are used. This may pose difficulties to implementing averaging over N angles in practice.} A more thorough analysis of which angle augmentations most effectively suppress the oscillations in the ISO coefficients is warranted. Here we simply choose angles which differ from the test angles ($2.5 \degree$ spacing) by multiples of $180\degree/32$ in order to be incommensurate with other frequencies present. The fractional RMS for $N=1,4,16$ is shown in Figure \ref{fig:CoeffStable} (bottom right) and decreases with larger $N$ as expected.

Increasing both $w$ and $N$ improve the angular stability of the ISO coefficients but have drawbacks. Increasing $N$ linearly increases the computational time spent computing scattering coefficients. Increasing $w$ also has an added computational cost since the sparsity of wavelets in Fourier space (which EqWS.jl takes advantage of) is decreased. Further, increasing $w$ decreases the sharpness of the angular response of EqWS (see Figure \ref{fig:RotEquiv}). However, because the wavelets are never constant and smoothly vary as a function of Fourier angular coordinate, this may not inhibit the ability of a classifier to pinpoint angular information, as hinted at in the MNIST rotation angle tests (Figure \ref{fig:LSeries}).

All of the representations here obscure the identity of each coefficient, but visualizing the RMS as a function of coefficient identity reveals only trends consistent with previous intuition about coefficient magnitudes. For $S_2(j_1,j_2,\ell_1.\ell_2)$, when $j_2 < j_1$ coefficients are small and, as a heuristic, the coefficients decrease with increasing $|j_1-j_2|$. Thus, while not a strict trend, coefficients with $j_2 < j_1$ have large fractional RMS fluctuations, but small magnitudes.

\subsection{Throw-Out Test}\label{sec:CoeffStabDropTest}

\begin{figure}[ht!]
\centering
\includegraphics[width=\linewidth]{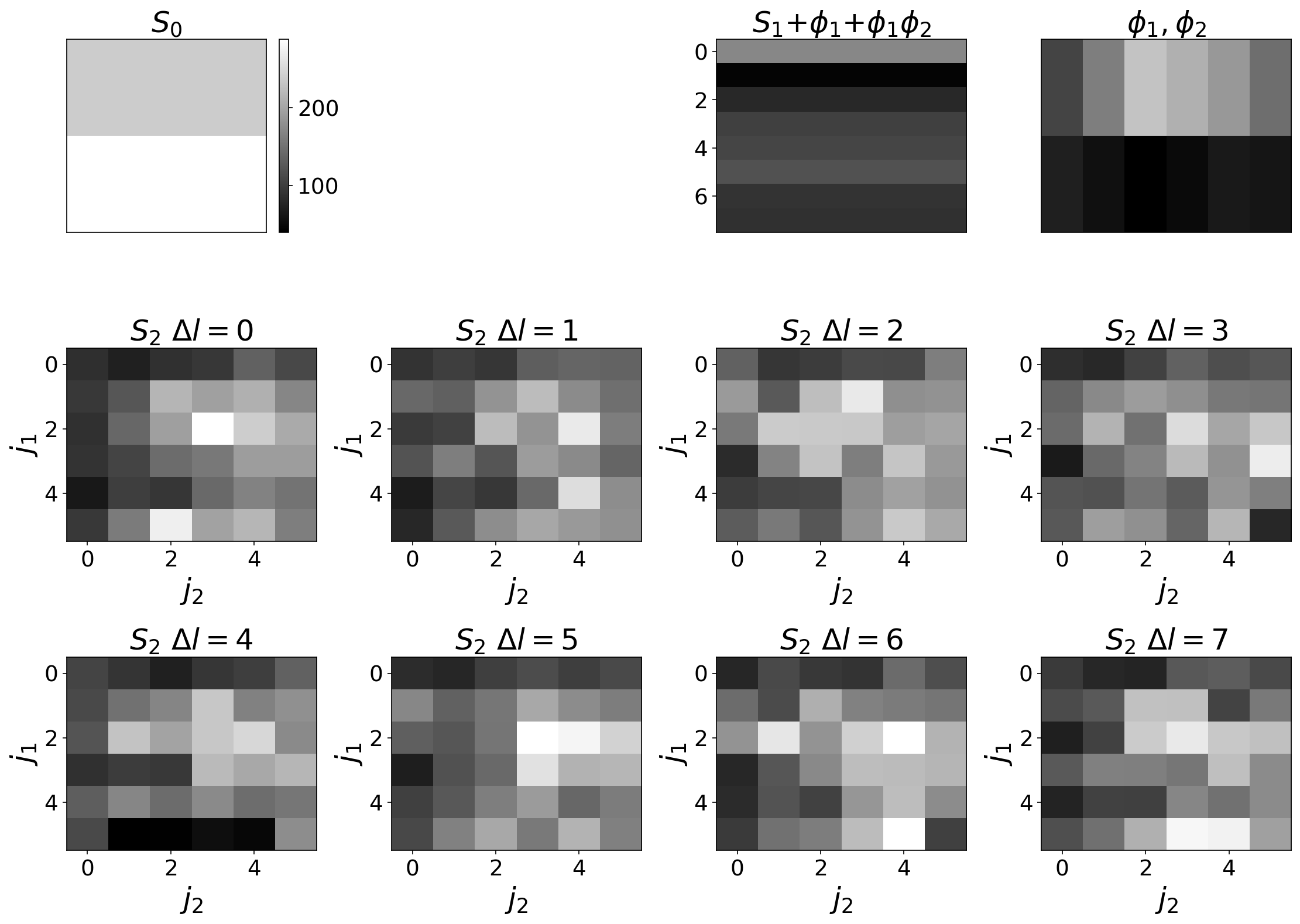}
\caption{The average index (from $10$ trials) at which a coefficient was discarded from Figure \ref{fig:MLQA} (bottom left) represented by the color scale. The global color-scale is indicated by the color bar of the $S_0$ subplot and ranges from global min to max. In the $S_0$ subplot, the top cell represents the average value of the image, and the bottom cell represents the image power. In the $S_1$ subplot, the first $6$ indices correspond to the $S_1$ coefficients $j \in [1, ..., 6]$. The last two indices correspond to the first-order $\phi$ term and second-order term involving $\phi$ at both steps. The top rightmost subplot contains the second-order $\psi$, $\phi$ terms. The top row depicts the second-order terms where $phi$ is used in the first layer, where the horizontal index corresponds to the $j$ index. The second row depicts the second-order terms where $phi$ is used in the second layer. The bottom $8$ panels correspond to second-order $\psi$-$\psi$ coefficients where each panel corresponds to a different $\Delta\ell$, and each pixel is indexed by $j_1$ and $j_2$ as indicated on the axes.
} 
\label{fig:DropOrder}
\end{figure}

We visualize the average order in which coefficients are thrown out during the throw-out test described in text and shown in Figure \ref{fig:MLQA}. We report the average step at which a coefficient was discarded, $1$ meaning the coefficient was discarded first in all $10$ trials and $310$ meaning the coefficient was the singular coefficient remaining at the end (Figure \ref{fig:DropOrder}). It is often useful to organize coefficient-dependent quantities by scales and angular divisions. While some circular representations have been used in the past \cite{Bruna:2012:arXiv:}, they can be difficult to interpret. We present the average discard index of the ISO coefficients instead in a matrix-like representation. The color scale ranges from the minimum to maximum index ($40, 288$). The index at which a coefficient was thrown out can be used in part as a proxy for how useful that coefficient is for the linear classifier used on this specific problem, MNIST. This analysis is further complicated by the fact that a linear combination of a set of coefficients might be the truly informative direction, and removing any one of those coefficients severely hinders classification. In addition, some coefficients which are not generally informative, but help the classifier overfit to the train data, may be retained longer. 

Despite these drawbacks, we observe general trends about how long coefficients are retained. We refer to wavelets with angular and scale indices as $\psi$ in contrast to the wavelet accounting for power at the origin, $\phi$. $S_0$ coefficients are retained longer than most other coefficients, with the $S_0$ coefficient associated with the image power lasting longer than that associated with the image mean. Only the $S_1(j=1)$ coefficient survives past the $50\%$ mark. The $j_1 = 3$ and nearby scale crossed $\phi$-$\psi$ terms are retained past the $50\%$ mark, but crossed $\psi$-$\phi$ terms are thrown out very quickly. For the S2 $\psi$-$\psi$ terms, we observe that intermediate scales with small $|j_1-j_2|$ are often retained the longest. The coefficients involving the smallest and largest scales are often thrown out earliest, as are those with $j_2 < j_1$, though neither is a strict rule. The lack of clear trends in the $S_2$ coefficients is troubling and suggests it may be worth revisiting the $j_2 > j_1$ restriction which is often made in scattering networks. 

\section{Extended Applications}\label{sec:ExtApp}

\subsection{MNIST Test-Train Matrix}\label{sec:MNISTTest-Train}

\begin{figure}[ht!]
\centering
\includegraphics[width=\linewidth]{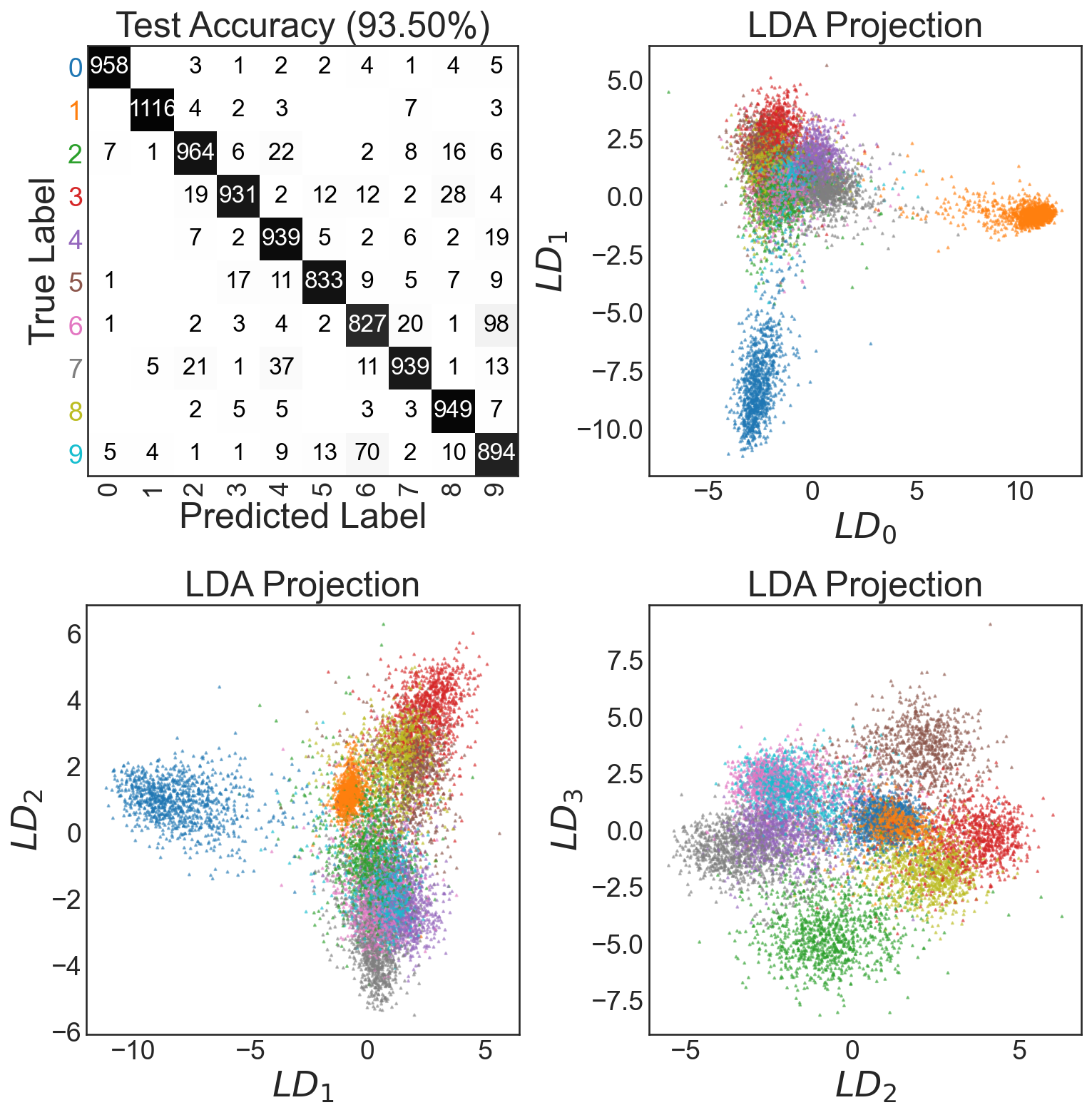}
\caption{Confusion matrix between MNIST classes when applying EqWS-ISO+LDA. Matrix entries are the number of test images with the corresponding true-predicted label pairing. The grayscale color indicates the true positive percentage relative to the number of test images with the corresponding true label. Colors correspond to true labels, as indicated on the vertical axis in the top left panel. Test images are shown in the first few components of LDA space.
}
\label{fig:MNISTtesttrain}
\end{figure}

We often find it instructive to inspect the test-train matrix and test feature vectors in the first few components of the LDA classification space. These are presented in Figure \ref{fig:MNISTtesttrain} for EqWS-ISO+LDA ($L=8$) on MNIST embedded in a $128 \times 128$ image interpolated to a $256 \times 256$ image. Both ``0'' and ``1'' are well separated by the first two LDA components. The largest confusion between ``6'' and ``9'', which is expected to increase using an isotropic representation, is also reflected in the LDA spaces where ``6'' and ``9'' overlap in $LD_2$ and $LD_3$ even when all of the other classes except ``0'' and ``1'' appear to have separated.

\subsection{Color Space Survey}\label{sec:ColorSpaceSurvey}

For color image classification, the color-space representation can alter the classification performance. We demonstrate the variability in classification for CIFAR-10 and CIFAR-100 obtained in $16$ different color spaces using EqWS-ISO + LDA. Preprocessing the EqWS coefficients by taking the logarithm and standard scaling is often used in practice to improve performance (``SC-log'') and is contrasted to using the raw coefficients (``None''). Preprocessing the coefficients and using YCbCr provided the best performance of the cases studied here.

\begin{table}[ht]
\centering
\caption{Accuracy of EqWS+LDA on CIFAR with Different Color Spaces}
\begin{tabular}{l|cc|cc}
\toprule

\multirow{2}{*}{Color Space} & \multicolumn{2}{c|}{CIFAR-10} & \multicolumn{2}{c}{CIFAR-100} \\
 & None & SC-log & None & SC-log \\

\midrule
RGB & $48.58$ & $50.27$ & $34.03$ & $35.63$ \\
YCbCr & $51.30$ &  \bm{$53.93$} & $35.02$ & \bm{$37.42$} \\
YIQ & $51.79$ & $53.08$ & $35.67$ & $36.99$ \\
HSV & $50.67$ & $52.22$ & $33.88$ & $34.44$ \\
HSL & $49.49$ & $51.50$ & $32.56$ & $33.24$ \\
HSI & $50.23$ & $52.55$  & $33.63$ & $35.16$ \\
Lab & $51.66$ & $52.54$ & $35.17$ & $35.62$ \\
Luv & \bm{$52.19$} & $53.38$ & $34.90$ & $36.30$ \\
LCHab & $50.98$ & $52.97$ & $34.63$ & $35.49$ \\
LCHuv & $51.06$ & $53.13$ & $34.36$ & $35.67$ \\
LMS & $45.55$ & $48.00$ & $31.52$ & $33.88$ \\
xyY & $47.95$ & $49.91$ & $31.89$ & $34.32$ \\
XYZ & $45.34$ & $47.95$ & $31.50$ & $33.96$ \\
DIN99 & $52.08$ & $53.34$  & $35.13$ & $35.20$ \\
DIN99d & $51.78$ & $52.76$ & $35.59$ & $36.22$ \\
DIN99o & $51.70$ & $53.24$ & \bm{$35.75$} & $36.02$ \\
\bottomrule
\end{tabular}
\end{table}

\subsection{Coefficient Reduction Comparison}\label{sec:ReducCompare}

\begin{figure}[t!]
\centering
\includegraphics[width=\linewidth]{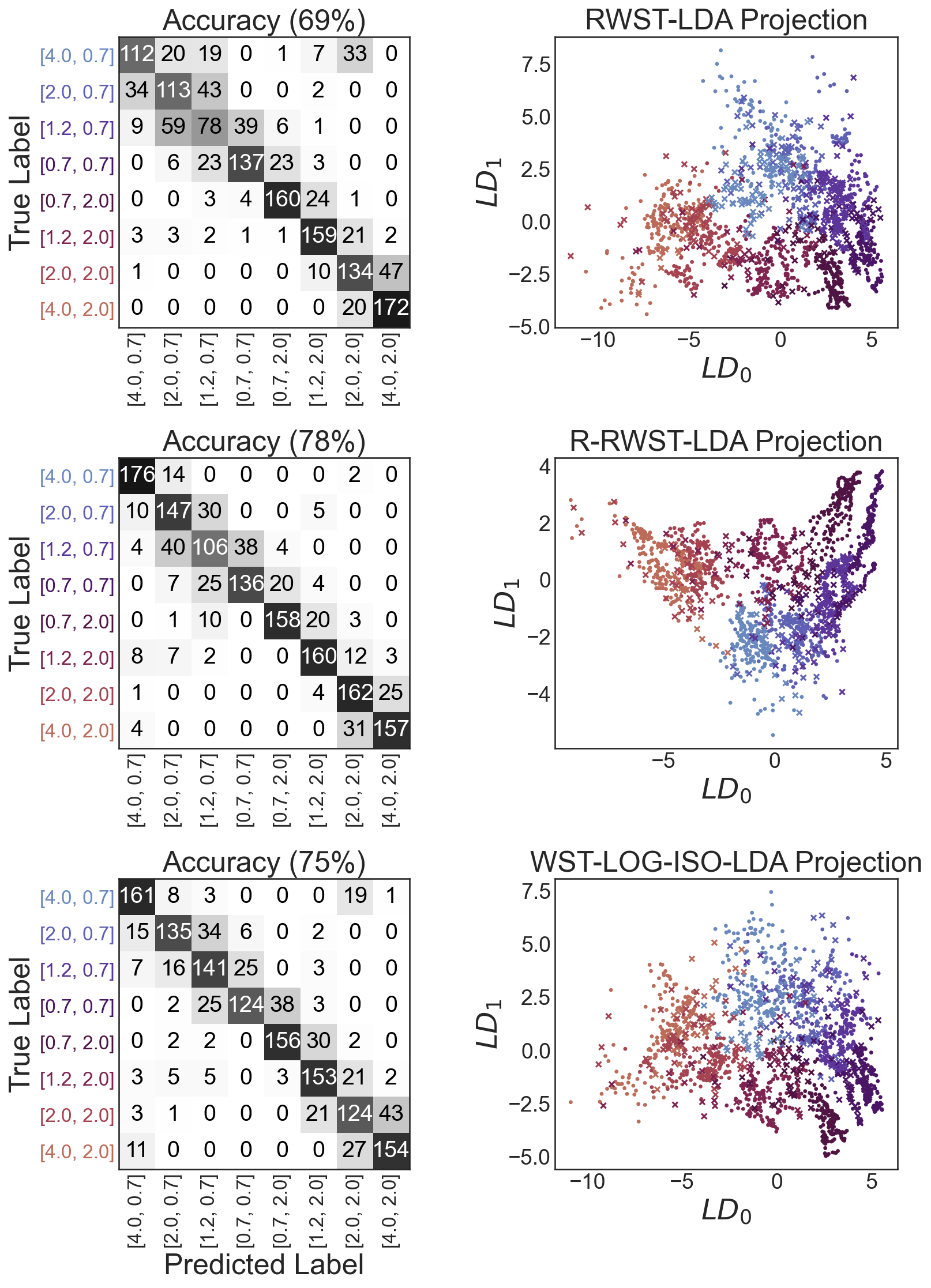}
\caption{Confusion matrix between MHD classes when applying RWST (top row), R-RWST (middle row), or WST-LOG-ISO (bottom row) combined with LDA to the 2D image classification. Matrix entries are the number of test images with the corresponding true-predicted label pairing. The grayscale color indicates the true positive percentage relative to the number of test images with the corresponding true label. Colors correspond to true labels, as indicated on the vertical axis in the top left panel. Test images are shown in the first few components of LDA space. The improved performance of R-RWST, using only the constant terms of the RWST fit, and WST-LOG-ISO, which is a sum over the angular index after taking the logarithm of the coefficients, suggests the isotropic coefficient reduction is sufficient for this task.
}
\label{fig:RWSTCompare}
\end{figure}

The first few components of the LDA space for classification of the MHD images are shown in Figure \ref{fig:RWSTCompare}. The reduced-wavelet scattering transform (RWST) to which we compare is defined on the original coefficients which are then normalized following \cite{Allys:2019:A&A:} and \cite{Bruna:2013:arXiv:}. 
\begin{align}\label{eq:RWSTNorm}
\bar S_0 & = \log_2\left[S_0\right] \nonumber\\
\bar S_1(j_1,\ell_1) & = \log_2\left[S_1(j_1,\ell_1)/S_0\right] \\ 
\bar S_2(j_1,\ell_1,j_2,\ell_2) & = \log_2\left[S_2(j_1,\ell_1,j_2,\ell_2)/S_1(j_1,\ell_1)\right] \nonumber
\end{align}
The RWST reduction takes advantage of periodicity and angular regularity observed in the WST coefficients to remove the angular indices. The RWST coefficients are obtained from the WST coefficients by least-squares fit of the first and second-order coefficients. For first order,
\begin{align}\label{eq:RWST1}
\bar S_1(j_1,\ell_1) = & S_1^{\rm iso}(j_1) + \nonumber\\
& S_1^{\rm aniso}(j_1) \cos\left[\frac{360\degree}{L}(\ell-\ell_{\rm a}^{\rm ref})\right]
\end{align}
where $S_1^{\rm iso}(j_1)$, $S_1^{\rm aniso}(j_1)$, and $\ell_{\rm a}^{\rm ref}(j_1)$ are fit coefficients. For second order, 
\begin{align}\label{eq:RWST2}
\bar S_2(j_1,\ell_1,j_2,\ell_2) = & S_2^{\rm iso,1}(j_1,j_2) + \\
& S_2^{\rm iso,2}(j_1,j_2) \cos\left[\frac{360\degree}{L}(\ell_1-\ell_2)\right] + \nonumber\\
& S_2^{\rm aniso,1}(j_1,j_2) \cos\left[\frac{360\degree}{L}(\ell_1-\ell_{\rm b}^{\rm ref})\right] + \nonumber\\
& S_2^{\rm aniso,2}(j_1,j_2) \cos\left[\frac{360\degree}{L}(\ell_2-\ell_{\rm b}^{\rm ref})\right] \nonumber
\end{align}
where $S_2^{\rm iso,1}(j_1,j_2)$, $S_2^{\rm iso,2}(j_1,j_2)$, $S_2^{\rm aniso,1}(j_1,j_2)$, $S_2^{\rm aniso,2}(j_1,j_2)$, and $\ell_{\rm b}^{\rm ref}(j_1,j_2)$ are fit coefficients. There are $3J$ coefficients for $m=1$, and $5J(J-1)/2$ coefficients for $m=2$, for a total of $J(5J+1)/2$.

We compare this to a simple isotropic average over the angular indices (WST-LOG-ISO), which is performed on the coefficients post normalization following Equation \ref{eq:RWSTNorm}. For first order this average removes the angular index entirely while for the second-order coefficients it leads to a dependence on only $\ell_1-\ell_2$. As a final comparison, we apply LDA only to the set of coefficients containing $\bar S_0$, $S_1^{\rm iso}(j_1)$, and $S_2^{\rm iso,1}(j_1,j_2)$, the constant terms which we refer to in Figure \ref{fig:RWSTCompare} as R-RWST.\footnote{Including $S_2^{\rm iso,2}$ in R-RWST decreases the accuracy to 72\%.}

The accuracy here is not directly comparable to the results obtained for RWST on MHD simulations in \cite{saydjari2020classification}. While we use the cumulative sum along the line-of-sight images from the same dataset, the images are standard scaled (set to have mean zero and standard deviation one) here instead of max-min $[0,1]$ scaled as in \cite{saydjari2020classification}. Standard scaling better separates learning the structure of fluctuations from learning the mean density/amplitude of fluctuations. The decrease in RWST performance on this task from $83\%$ in \cite{saydjari2020classification} to $69\%$ here suggests care must be taken in the choice of density-field normalization. While in applications, the mean density and amplitude of fluctuations can be informative and likely should be used, it is preferable to separate out this information when trying to compare different scattering networks, adding it back into the classifier in addition to scattering network coefficients if desired.

Apodization trades information which can be useful for classification in order to enable rotational invariant classifiers. While the best choice of this apodization function remains unresolved, we repeated the tests shown in Figure \ref{fig:RWSTCompare} with the simple apodization described in the main text. Only a small decrease in accuracy, $\sim 2\%$, was observed relative to Figure \ref{fig:RWSTCompare} with no apodization. This apodization allowed us to check that when either of the three cases presented in Figure \ref{fig:RWSTCompare} was trained at one angle and tested on rotated images, the classification accuracy was nearly random ($\sim18\%$). This result, as well as a similar one for testing on images with single pixel translations, in part motivated the development of EqWS.

\begin{figure}[t!]
\centering
\includegraphics[width=0.75\linewidth]{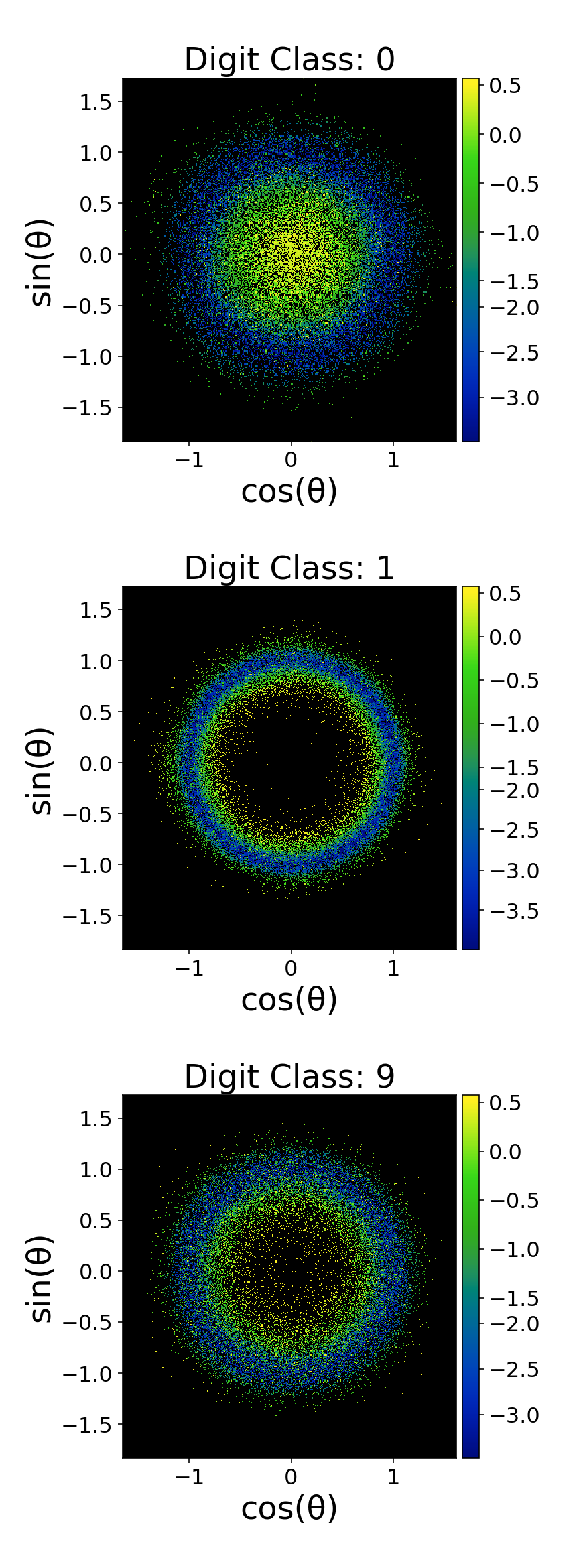}
\caption{A histogram of the predicted $\cos(\theta)$ and $\sin(\theta)$ by linear regression and colored by the logarithm of the squared error (SE) is shown for the extremal classes of $0$, $1$, and $9$. Here $L=4$ wavelets were used and the regression is performed where the class is known. Most images where the rotation angle was predicted with low error also fulfill $\sin^2(\theta) + \cos^2(\theta) = 1$ approximately. Note that the color bar is nonlinear even with respect to the log-SE because we use histogram equalization, which does not have fixed bin size. 
}
\label{fig:AngleRing}
\end{figure}

\subsection{MNIST Angle Estimation}\label{sec:MNISTAngles}

To estimate the angle of rotation for a given MNIST digit, we train a simple model which performs linear regression on $\cos(\theta)$ and $\sin(\theta)$, where $\theta$ is twice the rotation angle, and take the predicted $\theta$ to be the $\arctan$ of those components. However, one might worry that this model has no regularization fixing $\sin^2(\theta) + \cos^2(\theta) = 1$. We investigate the correlation between how closely the $\sin^2(\theta) + \cos^2(\theta) = 1$ is fulfilled and the sum of the squared error (SE) from fitting both the $\sin$ and $\cos$ components in Figure \ref{fig:AngleRing}. In each panel, a histogram colored by the logarithm of the SE is shown for a linear regression where the class is known and the extremal classes of $0$, $1$, and $9$ are shown. The wavelet filter bank here used $L=4$ and the same resizing and rotation scheme as Figure \ref{fig:LSeries}. Note that the color bar is nonlinear even with respect to the log-SE because we use histogram equalization, which does not have fixed bin size. Figure \ref{fig:AngleRing} illustrates that the $\sin^2(\theta) + \cos^2(\theta) = 1$ condition is approximately met for images where the rotation angle was estimated with low loss, but not for images with high loss. This suggests that the sum of squares of the $\sin$ and $\cos$ components could be used to estimate confidence in an angle prediction in a setting where the true labels were not known.

\ifCLASSOPTIONcaptionsoff
  \newpage
\fi

\bibliographystyle{IEEEtran}
\bibliography{equiv}

\begin{thebibliography}{10}
\providecommand{\url}[1]{#1}
\csname url@samestyle\endcsname
\providecommand{\newblock}{\relax}
\providecommand{\bibinfo}[2]{#2}
\providecommand{\BIBentrySTDinterwordspacing}{\spaceskip=0pt\relax}
\providecommand{\BIBentryALTinterwordstretchfactor}{4}
\providecommand{\BIBentryALTinterwordspacing}{\spaceskip=\fontdimen2\font plus
\BIBentryALTinterwordstretchfactor\fontdimen3\font minus
  \fontdimen4\font\relax}
\providecommand{\BIBforeignlanguage}[2]{{%
\expandafter\ifx\csname l@#1\endcsname\relax
\typeout{** WARNING: IEEEtran.bst: No hyphenation pattern has been}%
\typeout{** loaded for the language `#1'. Using the pattern for}%
\typeout{** the default language instead.}%
\else
\language=\csname l@#1\endcsname
\fi
#2}}
\providecommand{\BIBdecl}{\relax}
\BIBdecl

\bibitem{PlanckCollaboration:2017:A&A:}
{Planck Collaboration}, N.~{Aghanim} \emph{et~al.}, ``{Planck intermediate
  results. LI. Features in the cosmic microwave background temperature power
  spectrum and shifts in cosmological parameters},'' \emph{A\&A}, vol. 607, p.
  A95, Nov. 2017.

\bibitem{Peek:2019:ApJL:}
J.~E.~G. {Peek} and B.~{Burkhart}, ``{Do Androids Dream of Magnetic Fields?
  Using Neural Networks to Interpret the Turbulent Interstellar Medium},''
  \emph{ApJL}, vol. 882, no.~1, p. L12, Sep. 2019.

\bibitem{Peebles:2001:ASPC:}
P.~J.~E. {Peebles}, ``{The Galaxy and Mass N-Point Correlation Functions: a
  Blast from the Past},'' in \emph{Historical Development of Modern Cosmology},
  ser. Astronomical Society of the Pacific Conference Series, V.~J.
  {Mart{\'\i}nez}, V.~{Trimble}, and M.~J. {Pons-Border{\'\i}a}, Eds., vol.
  252, Jan. 2001, p. 201.

\bibitem{Burkhart:2009:ApJ:}
B.~{Burkhart}, D.~{Falceta-Gon{\c{c}}alves} \emph{et~al.}, ``{Density Studies
  of MHD Interstellar Turbulence: Statistical Moments, Correlations and
  Bispectrum},'' \emph{ApJ}, vol. 693, no.~1, pp. 250--266, Mar. 2009.

\bibitem{Burkhart:2016:ApJ:}
B.~{Burkhart} and A.~{Lazarian}, ``{The Phase Coherence of Interstellar Density
  Fluctuations},'' \emph{ApJ}, vol. 827, no.~1, p.~26, Aug. 2016.

\bibitem{zeiler2014visualizing}
M.~D. Zeiler and R.~Fergus, ``Visualizing and understanding convolutional
  networks,'' in \emph{European conference on computer vision}.\hskip 1em plus
  0.5em minus 0.4em\relax Springer, 2014, pp. 818--833.

\bibitem{zhou2014object}
B.~Zhou, A.~Khosla \emph{et~al.}, ``Object detectors emerge in deep scene
  cnns,'' \emph{arXiv preprint arXiv:1412.6856}, 2014.

\bibitem{karpathy2015visualizing}
A.~Karpathy, J.~Johnson, and L.~Fei-Fei, ``Visualizing and understanding
  recurrent networks,'' \emph{arXiv preprint arXiv:1506.02078}, 2015.

\bibitem{olah2017feature}
C.~Olah, A.~Mordvintsev, and L.~Schubert, ``Feature visualization,''
  \emph{Distill}, 2017, https://distill.pub/2017/feature-visualization.

\bibitem{bau2017network}
D.~Bau, B.~Zhou \emph{et~al.}, ``Network dissection: Quantifying
  interpretability of deep visual representations,'' in \emph{Proceedings of
  the IEEE conference on computer vision and pattern recognition}, 2017, pp.
  6541--6549.

\bibitem{cammarata:2020:curve}
N.~Cammarata, G.~Goh \emph{et~al.}, ``Curve detectors,'' \emph{Distill}, 2020,
  https://distill.pub/2020/circuits/curve-detectors.

\bibitem{Bruna:2012:arXiv:}
J.~{Bruna} and S.~{Mallat}, ``{Invariant Scattering Convolution Networks},''
  \emph{arXiv e-prints}, p. arXiv:1203.1513, Mar. 2012.

\bibitem{Mallat:2011:arXiv:}
S.~Mallat, ``Group invariant scattering,'' \emph{Communications on Pure and
  Applied Mathematics}, vol.~65, no.~10, pp. 1331--1398, 2012.

\bibitem{Hirn:2016:arXiv:}
M.~{Hirn}, S.~{Mallat}, and N.~{Poilvert}, ``{Wavelet Scattering Regression of
  Quantum Chemical Energies},'' \emph{arXiv e-prints}, p. arXiv:1605.04654, May
  2016.

\bibitem{Eickenberg:2018:JChPh:}
M.~{Eickenberg}, G.~{Exarchakis} \emph{et~al.}, ``{Solid harmonic wavelet
  scattering for predictions of molecule properties},'' \emph{JChPh}, vol. 148,
  no.~24, p. 241732, Jun. 2018.

\bibitem{Bruna:2013:arXiv:}
J.~{Bruna}, S.~{Mallat} \emph{et~al.}, ``{Intermittent process analysis with
  scattering moments},'' \emph{arXiv e-prints}, p. arXiv:1311.4104, Nov. 2013.

\bibitem{Allys:2019:A&A:}
E.~{Allys}, F.~{Levrier} \emph{et~al.}, ``{The RWST, a comprehensive
  statistical description of the non-Gaussian structures in the ISM},''
  \emph{A\&A}, vol. 629, p. A115, Sep. 2019.

\bibitem{saydjari2020classification}
A.~K. Saydjari, S.~K. Portillo \emph{et~al.}, ``Classification of
  magnetohydrodynamic simulations using wavelet scattering transforms,''
  \emph{arXiv preprint arXiv:2010.11963}, 2020.

\bibitem{Regaldo-SaintBlancard:2020:arXiv:}
B.~{Regaldo-Saint Blancard}, F.~{Levrier} \emph{et~al.}, ``{Statistical
  description of dust polarized emission from the diffuse interstellar medium
  -- A RWST approach},'' \emph{arXiv e-prints}, p. arXiv:2007.08242, Jul. 2020.

\bibitem{Allys:2020:arXiv:}
E.~{Allys}, T.~{Marchand} \emph{et~al.}, ``{New Interpretable Statistics for
  Large Scale Structure Analysis and Generation},'' \emph{arXiv e-prints}, p.
  arXiv:2006.06298, Jun. 2020.

\bibitem{villaescusa2020quijote}
F.~Villaescusa-Navarro, C.~Hahn \emph{et~al.}, ``The quijote simulations,''
  \emph{The Astrophysical Journal Supplement Series}, vol. 250, no.~1, p.~2,
  2020.

\bibitem{Cheng:2020:arXiv:}
S.~{Cheng}, Y.-S. {Ting} \emph{et~al.}, ``{A new approach to observational
  cosmology using the scattering transform},'' \emph{arXiv e-prints}, p.
  arXiv:2006.08561, Jun. 2020.

\bibitem{Cheng:2021:arXiv:}
S.~{Cheng} and B.~{M{\'e}nard}, ``{Weak lensing scattering transform: dark
  energy and neutrino mass sensitivity},'' \emph{arXiv e-prints}, p.
  arXiv:2103.09247, Mar. 2021.

\bibitem{Angles:2018:arXiv:}
T.~{Angles} and S.~{Mallat}, ``{Generative networks as inverse problems with
  Scattering transforms},'' \emph{arXiv e-prints}, p. arXiv:1805.06621, May
  2018.

\bibitem{Bruna:2018:arXiv:}
J.~{Bruna} and S.~{Mallat}, ``{Multiscale Sparse Microcanonical Models},''
  \emph{arXiv e-prints}, p. arXiv:1801.02013, Jan. 2018.

\bibitem{Regaldo-SaintBlancard:2021:arXiv:}
B.~{Regaldo-Saint Blancard}, E.~{Allys} \emph{et~al.}, ``{A new approach for
  the statistical denoising of Planck interstellar dust polarization data},''
  \emph{arXiv e-prints}, p. arXiv:2102.03160, Feb. 2021.

\bibitem{engstrom2019exploring}
L.~Engstrom, B.~Tran \emph{et~al.}, ``Exploring the landscape of spatial
  robustness,'' in \emph{International Conference on Machine Learning}.\hskip
  1em plus 0.5em minus 0.4em\relax PMLR, 2019, pp. 1802--1811.

\bibitem{scherer2010evaluation}
D.~Scherer, A.~M{\"u}ller, and S.~Behnke, ``Evaluation of pooling operations in
  convolutional architectures for object recognition,'' in \emph{International
  conference on artificial neural networks}.\hskip 1em plus 0.5em minus
  0.4em\relax Springer, 2010, pp. 92--101.

\bibitem{azulay2018deep}
A.~Azulay and Y.~Weiss, ``Why do deep convolutional networks generalize so
  poorly to small image transformations?'' \emph{arXiv preprint
  arXiv:1805.12177}, 2018.

\bibitem{zhang2019making}
R.~Zhang, ``Making convolutional networks shift-invariant again,'' in
  \emph{International Conference on Machine Learning}.\hskip 1em plus 0.5em
  minus 0.4em\relax PMLR, 2019, pp. 7324--7334.

\bibitem{lecun1998gradient}
Y.~LeCun, L.~Bottou \emph{et~al.}, ``Gradient-based learning applied to
  document recognition,'' \emph{Proceedings of the IEEE}, vol.~86, no.~11, pp.
  2278--2324, 1998.

\bibitem{nyquist1928certain}
H.~Nyquist, ``Certain topics in telegraph transmission theory,''
  \emph{Transactions of the American Institute of Electrical Engineers},
  vol.~47, no.~2, pp. 617--644, 1928.

\bibitem{jacobsen2016structured}
J.-H. Jacobsen, J.~Van~Gemert \emph{et~al.}, ``Structured receptive fields in
  cnns,'' in \emph{Proceedings of the IEEE Conference on Computer Vision and
  Pattern Recognition}, 2016, pp. 2610--2619.

\bibitem{laptev2016ti}
D.~Laptev, N.~Savinov \emph{et~al.}, ``Ti-pooling: transformation-invariant
  pooling for feature learning in convolutional neural networks,'' in
  \emph{Proceedings of the IEEE conference on computer vision and pattern
  recognition}, 2016, pp. 289--297.

\bibitem{cohen2016group}
T.~Cohen and M.~Welling, ``Group equivariant convolutional networks,'' in
  \emph{International conference on machine learning}.\hskip 1em plus 0.5em
  minus 0.4em\relax PMLR, 2016, pp. 2990--2999.

\bibitem{cohen2016steerable}
T.~S. Cohen and M.~Welling, ``Steerable cnns,'' \emph{arXiv preprint
  arXiv:1612.08498}, 2016.

\bibitem{weiler2019general}
M.~Weiler and G.~Cesa, ``General $ e (2) $-equivariant steerable cnns,''
  \emph{arXiv preprint arXiv:1911.08251}, 2019.

\bibitem{romero2019co}
D.~W. Romero and M.~Hoogendoorn, ``Co-attentive equivariant neural networks:
  Focusing equivariance on transformations co-occurring in data,'' \emph{arXiv
  preprint arXiv:1911.07849}, 2019.

\bibitem{romero2020attentive}
D.~Romero, E.~Bekkers \emph{et~al.}, ``Attentive group equivariant
  convolutional networks,'' in \emph{International Conference on Machine
  Learning}.\hskip 1em plus 0.5em minus 0.4em\relax PMLR, 2020, pp. 8188--8199.

\bibitem{lafarge2021roto}
M.~W. Lafarge, E.~J. Bekkers \emph{et~al.}, ``Roto-translation equivariant
  convolutional networks: Application to histopathology image analysis,''
  \emph{Medical Image Analysis}, vol.~68, p. 101849, 2021.

\bibitem{cohen2018spherical}
T.~S. Cohen, M.~Geiger \emph{et~al.}, ``Spherical cnns,'' \emph{arXiv preprint
  arXiv:1801.10130}, 2018.

\bibitem{kondor2018clebsch}
R.~Kondor, Z.~Lin, and S.~Trivedi, ``Clebsch-gordan nets: a fully fourier space
  spherical convolutional neural network,'' \emph{arXiv preprint
  arXiv:1806.09231}, 2018.

\bibitem{esteves2018learning}
C.~Esteves, C.~Allen-Blanchette \emph{et~al.}, ``Learning so (3) equivariant
  representations with spherical cnns,'' in \emph{Proceedings of the European
  Conference on Computer Vision (ECCV)}, 2018, pp. 52--68.

\bibitem{cohen2019gauge}
T.~Cohen, M.~Weiler \emph{et~al.}, ``Gauge equivariant convolutional networks
  and the icosahedral cnn,'' in \emph{International Conference on Machine
  Learning}.\hskip 1em plus 0.5em minus 0.4em\relax PMLR, 2019, pp. 1321--1330.

\bibitem{defferrard2020deepsphere}
M.~Defferrard, M.~Milani \emph{et~al.}, ``Deepsphere: a graph-based spherical
  cnn,'' \emph{arXiv preprint arXiv:2012.15000}, 2020.

\bibitem{esteves:2020:spin}
C.~Esteves, A.~Makadia, and K.~Daniilidis, ``Spin-weighted spherical cnns,''
  \emph{arXiv preprint arXiv:2006.10731}, 2020.

\bibitem{sifre:2013:rotation}
L.~Sifre and S.~Mallat, ``Rotation, scaling and deformation invariant
  scattering for texture discrimination,'' in \emph{Proceedings of the IEEE
  conference on computer vision and pattern recognition}, 2013, pp. 1233--1240.

\bibitem{worrall2019deep}
D.~E. Worrall and M.~Welling, ``Deep scale-spaces: Equivariance over scale,''
  \emph{arXiv preprint arXiv:1905.11697}, 2019.

\bibitem{sosnovik2019scale}
I.~Sosnovik, M.~Szmaja, and A.~Smeulders, ``Scale-equivariant steerable
  networks,'' \emph{arXiv preprint arXiv:1910.11093}, 2019.

\bibitem{romero2020wavelet}
D.~W. Romero, E.~J. Bekkers \emph{et~al.}, ``Wavelet networks: Scale
  equivariant learning from raw waveforms,'' \emph{arXiv preprint
  arXiv:2006.05259}, 2020.

\bibitem{ma2010curvelet}
J.~Ma and G.~Plonka, ``The curvelet transform,'' \emph{IEEE signal processing
  magazine}, vol.~27, no.~2, pp. 118--133, 2010.

\bibitem{Andreux:2018:arXiv:}
M.~{Andreux}, T.~{Angles} \emph{et~al.}, ``{Kymatio: Scattering Transforms in
  Python},'' \emph{arXiv e-prints}, p. arXiv:1812.11214, Dec. 2018.

\bibitem{selesnick:2005:dual}
I.~W. Selesnick, R.~G. Baraniuk, and N.~C. Kingsbury, ``The dual-tree complex
  wavelet transform,'' \emph{IEEE signal processing magazine}, vol.~22, no.~6,
  pp. 123--151, 2005.

\bibitem{cohen:2018:spherical}
T.~S. Cohen, M.~Geiger \emph{et~al.}, ``Spherical cnns,'' \emph{arXiv preprint
  arXiv:1801.10130}, 2018.

\bibitem{kondor:2018:clebsch}
R.~Kondor, Z.~Lin, and S.~Trivedi, ``Clebsch-gordan nets: a fully fourier space
  spherical convolutional neural network,'' \emph{arXiv preprint
  arXiv:1806.09231}, 2018.

\bibitem{blankrot:rotmnist}
B.~Blankrot, ``Detecting rotation of handwritten numbers,''
  \url{https://github.com/bblankrot/detect_number_rot}, 2021.

\bibitem{cohen:2017:emnist}
G.~Cohen, S.~Afshar \emph{et~al.}, ``Emnist: Extending mnist to handwritten
  letters,'' in \emph{2017 International Joint Conference on Neural Networks
  (IJCNN)}.\hskip 1em plus 0.5em minus 0.4em\relax IEEE, 2017, pp. 2921--2926.

\bibitem{krizhevsky:2009:learning}
A.~Krizhevsky, G.~Hinton \emph{et~al.}, ``Learning multiple layers of features
  from tiny images,'' 2009.

\bibitem{oyallon2015deep}
E.~Oyallon and S.~Mallat, ``Deep roto-translation scattering for object
  classification,'' in \emph{Proceedings of the IEEE Conference on Computer
  Vision and Pattern Recognition}, 2015, pp. 2865--2873.

\bibitem{cotter2017visualizing}
F.~Cotter and N.~Kingsbury, ``Visualizing and improving scattering networks,''
  in \emph{2017 IEEE 27th International Workshop on Machine Learning for Signal
  Processing (MLSP)}.\hskip 1em plus 0.5em minus 0.4em\relax IEEE, 2017, pp.
  1--6.

\bibitem{collette_python_hdf5_2014}
A.~Collette, \emph{Python and HDF5}.\hskip 1em plus 0.5em minus 0.4em\relax
  O'Reilly, 2013.

\bibitem{Perez:2007:CSE:}
F.~{Perez} and B.~E. {Granger}, ``{IPython: A System for Interactive Scientific
  Computing},'' \emph{Computing in Science and Engineering}, vol.~9, no.~3, pp.
  21--29, Jan. 2007.

\bibitem{Hunter:2007:CSE:}
J.~D. {Hunter}, ``{Matplotlib: A 2D Graphics Environment},'' \emph{Computing in
  Science and Engineering}, vol.~9, no.~3, pp. 90--95, May 2007.

\bibitem{vanderWalt:2011:CSE:}
S.~{van der Walt}, S.~C. {Colbert}, and G.~{Varoquaux}, ``{The NumPy Array: A
  Structure for Efficient Numerical Computation},'' \emph{Computing in Science
  and Engineering}, vol.~13, no.~2, pp. 22--30, Mar. 2011.

\bibitem{Virtanen:2020:NatMe:}
P.~{Virtanen}, R.~{Gommers} \emph{et~al.}, ``{SciPy 1.0: fundamental algorithms
  for scientific computing in Python},'' \emph{Nature Methods}, vol.~17, pp.
  261--272, Feb. 2020.

\bibitem{Pedregosa:2012:arXiv:}
F.~{Pedregosa}, G.~{Varoquaux} \emph{et~al.}, ``{Scikit-learn: Machine Learning
  in Python},'' \emph{arXiv e-prints}, p. arXiv:1201.0490, Jan. 2012.

\bibitem{vanderWalt:2014:arXiv:}
S.~{van der Walt}, J.~L. {Sch{\"o}nberger} \emph{et~al.}, ``{scikit-image:
  Image processing in Python},'' \emph{arXiv e-prints}, p. arXiv:1407.6245,
  Jul. 2014.

\bibitem{Speagle:2020:MNRAS:}
J.~S. {Speagle}, ``{DYNESTY: a dynamic nested sampling package for estimating
  Bayesian posteriors and evidences},'' \emph{MNRAS}, vol. 493, no.~3, pp.
  3132--3158, Apr. 2020.

\bibitem{bezanson2017julia}
\BIBentryALTinterwordspacing
J.~Bezanson, A.~Edelman \emph{et~al.}, ``Julia: A fresh approach to numerical
  computing,'' \emph{SIAM review}, vol.~59, no.~1, pp. 65--98, 2017. [Online].
  Available: \url{https://doi.org/10.1137/141000671}
\BIBentrySTDinterwordspacing

\bibitem{FFTW.jl-2005}
M.~Frigo and S.~G. Johnson, ``The design and implementation of {FFTW3},''
  \emph{Proceedings of the IEEE}, vol.~93, no.~2, pp. 216--231, 2005, special
  issue on ``Program Generation, Optimization, and Platform Adaptation''.

\bibitem{mallat2020phase}
S.~Mallat, S.~Zhang, and G.~Rochette, ``Phase harmonic correlations and
  convolutional neural networks,'' \emph{Information and Inference: A Journal
  of the IMA}, vol.~9, no.~3, pp. 721--747, 2020.

\bibitem{Ashmead:2010:arXiv:}
J.~{Ashmead}, ``{Morlet wavelets in quantum mechanics},'' \emph{arXiv
  e-prints}, p. arXiv:1001.0250, Jan. 2010.

\end{thebibliography}


\begin{IEEEbiography}[{\includegraphics[width=1in,height=1.25in,clip,keepaspectratio,trim=0cm -5cm 0 0]{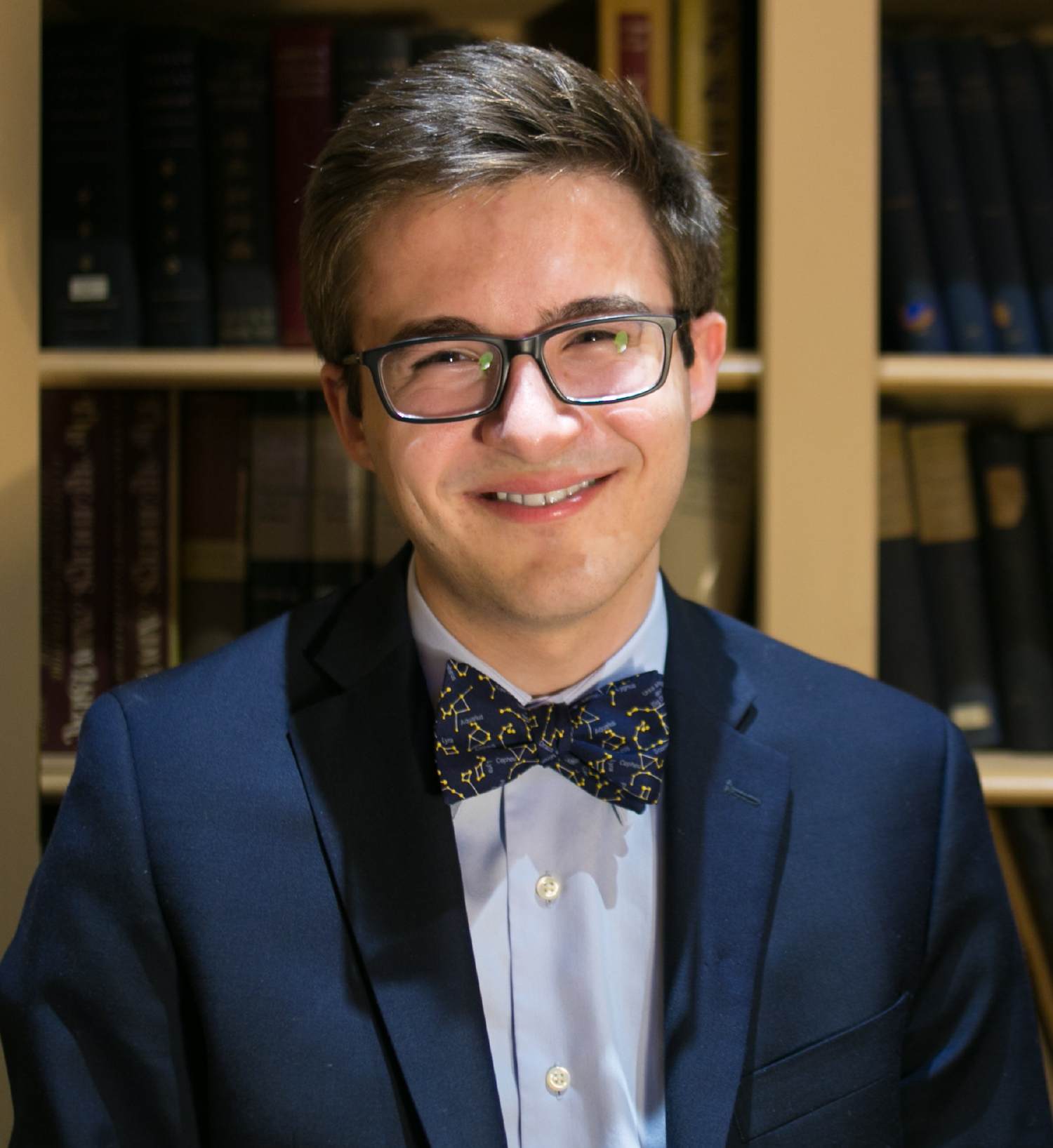}}]{Andrew K. Saydjari}
received a B.Sc. in mathematics and B.Sc./M.Sc in chemistry from Yale in 2018. He is currently pursuing a Ph.D. in Physics at Harvard as an NSF Graduate Research Fellow.
His research interests include equivariant machine learning and interpretable yet robust image statistics. His current applications of choice are non-Gaussian astrophysical processes such as interstellar dust.
\end{IEEEbiography}

\begin{IEEEbiography}[{\includegraphics[width=1in,height=1.25in,clip,keepaspectratio,trim=0cm -5cm 0 0]{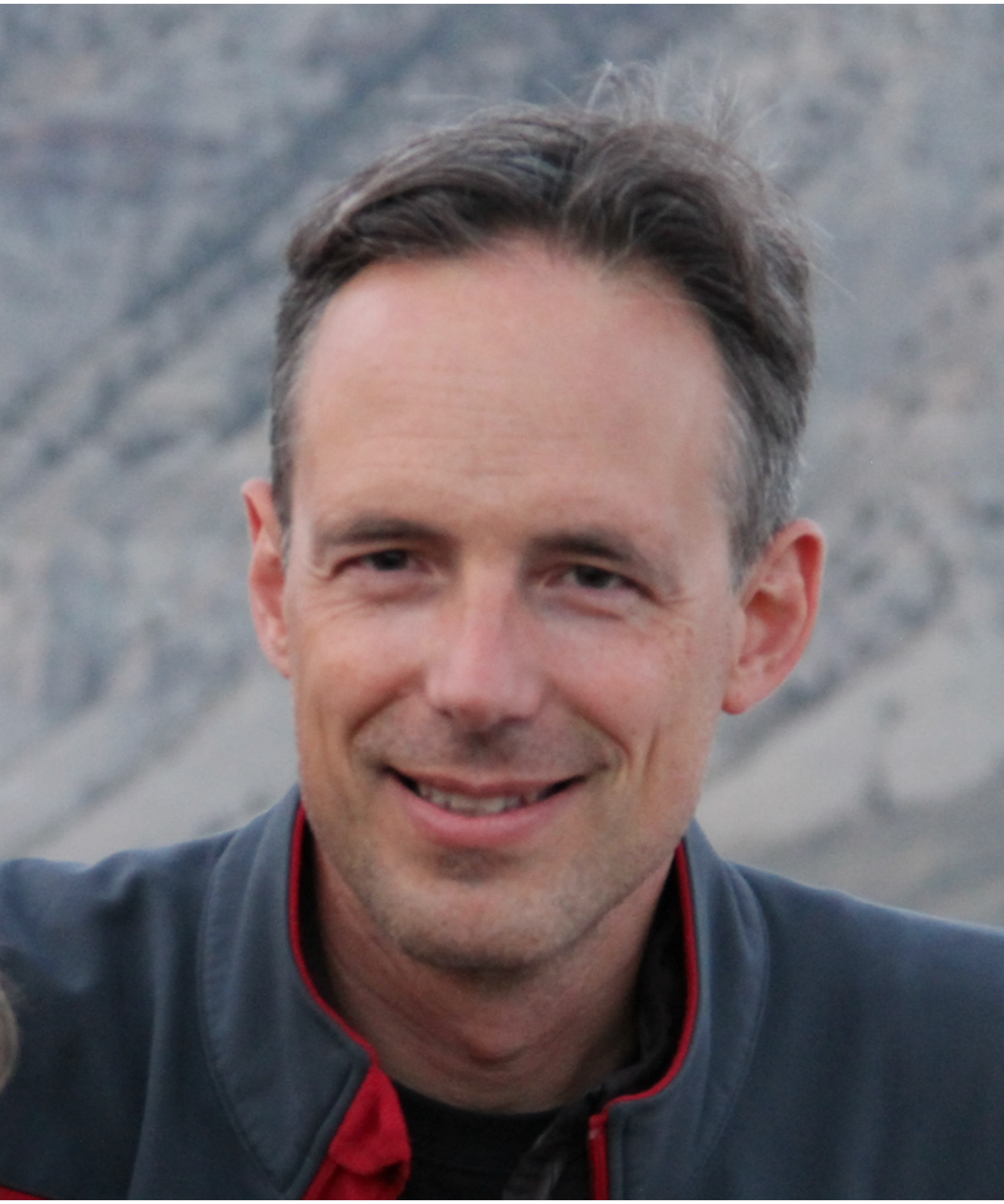}}]{Douglas P. Finkbeiner}
received his Ph.D. in Physics from UC Berkeley in 1999. He is currently a professor of Astronomy and of Physics at Harvard University. His research involves inference problems in large astronomical datasets, including mapping interstellar dust in 2 and 3 dimensions using the colors of billions of stars. 
\end{IEEEbiography}

\end{document}